\documentclass[10pt,journal,compsoc]{IEEEtran}

\ifCLASSOPTIONcompsoc

  \usepackage[nocompress]{cite}
\else

  \usepackage{cite}
\fi
\usepackage{amsmath}
\usepackage{amsfonts}
\usepackage{url}
\usepackage{graphicx}
\usepackage{booktabs}
\usepackage{color}
\usepackage{colortbl}
\usepackage{xcolor}

\graphicspath{{Figures-PDF/}}

\ifCLASSINFOpdf
  
\else
  
\fi

\begin{document}

\title{Memory-Guided Collaborative Attention for Nighttime Thermal Infrared Image Colorization}

\author{Fu-Ya~Luo,
        Yi-Jun~Cao,
        Kai-Fu~Yang,
        and~Yong-Jie~Li,~\IEEEmembership{Senior~Member,~IEEE}% 
\IEEEcompsocitemizethanks{\IEEEcompsocthanksitem Fu-Ya Luo, Yi-Jun Cao, Kai-Fu Yang and Yong-Jie Li 
are with the MOE Key Laboratory for Neuroinformation, the School of Life Science and Tecnology, 
University of Electronic Science and Technology of China, Chengdu 610054, 
China.(\textit{Corresponding author: Yong-Jie Li.})\protect\\

E-mail: luofuya1993@gmail.com, yijuncaoo@gmail.com, yangkf@uestc.edu.cn, 
liyj@uestc.edu.cn.}
}
% \thanks{Manuscript received April 19, 2005; revised August 26, 2015.}}

\markboth{Journal of \LaTeX\ Class Files,~Vol.~14, No.~8, August~2015}%
{Shell \MakeLowercase{\textit{et al.}}: Bare Demo of IEEEtran.cls for Computer Society Journals}

\IEEEtitleabstractindextext{%
\begin{abstract}
  Nighttime thermal infrared (NTIR) image colorization, also known as translation of NTIR 
  images into daytime color images (NTIR2DC), is a promising research direction to facilitate 
  nighttime scene perception for humans and intelligent systems under unfavorable conditions (e.g., 
  complete darkness). However, previously developed methods have poor colorization performance for small 
  sample classes. Moreover, reducing the high confidence noise in pseudo-labels and addressing 
  the problem of image gradient disappearance during translation are still under-explored, and 
  keeping edges from being distorted during translation is also challenging. To address the 
  aforementioned issues, we propose a novel learning framework called Memory-guided cOllaboRative 
  atteNtion Generative Adversarial Network (MornGAN), which is inspired by the analogical 
  reasoning mechanisms of humans. Specifically, a 
  memory-guided sample selection strategy and adaptive collaborative attention loss are 
  devised to enhance the semantic preservation of small sample categories. 
  In addition, we propose an online semantic distillation module to mine and refine the 
  pseudo-labels of NTIR images. Further, conditional gradient repair loss is introduced for 
  reducing edge distortion during translation. Extensive experiments on the NTIR2DC task 
  show that the proposed MornGAN significantly outperforms other image-to-image translation 
  methods in terms of semantic preservation and edge consistency, which helps improve the 
  object detection accuracy remarkably.
\end{abstract}

\begin{IEEEkeywords}
  Thermal infrared image colorization, image-to-image translation, generative adversarial 
  networks, memory-guided collaborative attention, nighttime scene perception.
\end{IEEEkeywords}}

\maketitle

\IEEEdisplaynontitleabstractindextext

\IEEEpeerreviewmaketitle

\IEEEraisesectionheading{\section{Introduction}\label{sec:introduction}}

\IEEEPARstart{A}{utomatic} driving and assisted driving systems need to ensure reliable 
all-weather scene perception, especially in unfavorable environments with, for example, 
nighttime low-light and daytime rain. 
Compared with light-sensitive visible spectrum-based sensors, thermal infrared (TIR) cameras, which 
can image in complete darkness and have high penetration in foggy environments, may be more 
suitable for all-weather scene perception. However, TIR images usually have low contrast and 
ambiguous object boundaries. 
In addition, the monochromatic nature of TIR images is not conducive to human 
interpretation \cite{2009-EOJ-W} and domain adaptation from RGB-based algorithms. Therefore, it is significant 
to translate nighttime TIR (NTIR) images into corresponding daytime color (DC) images, 
which can not only help drivers quickly perceive their surroundings in night conditions, 
but also reduce the annotation cost of NTIR image-understanding tasks by using 
existing annotated DC datasets. In this study, we explore NTIR image colorization, 
which is also called translation from NTIR to DC images (abbreviated as NTIR2DC).

Since vast quantities of pixel-level registered NTIR and DC image pairs are difficult to acquire, 
a potential solution for the NTIR2DC task is to utilize unpaired image-to-image (I2I) 
translation methods. Driven by the success of generative adversarial 
networks (GANs) \cite{2014-NIPS-Goodfellow} in high quality image generation, numerous studies 
have leveraged GANs to implement unpaired I2I translation \cite{2017-CVPR-Zhu,2017-NIPS-Liu}. 
Despite the impressive results, unpaired I2I translation methods frequently suffer 
from content distortion due to the lack of explicit semantic supervision. To mitigate this 
limitation, many efforts have been dedicated to introducing semantic consistency constraints using 
segmentation labels. For example, AugGAN \cite{2018-ECCV-Huang-Auggan} and 
Sem-GAN \cite{2019-WACV-Cherian} introduced additional segmentation branches to enforce the 
segmentation masks of the translated images to be consistent with the labels. In the 
case where only the semantic annotation of the source domain is available, 
\cite{2020-Arxiv-Musto} and \cite{2020-WACV-Pizzati} combined self-supervised learning 
and thresholding to generate pseudo-labels for the target domain images, which in turn 
constrain the semantically invariant image translation. 

\begin{figure}[!t]
\centering
\includegraphics[width=3.45in]{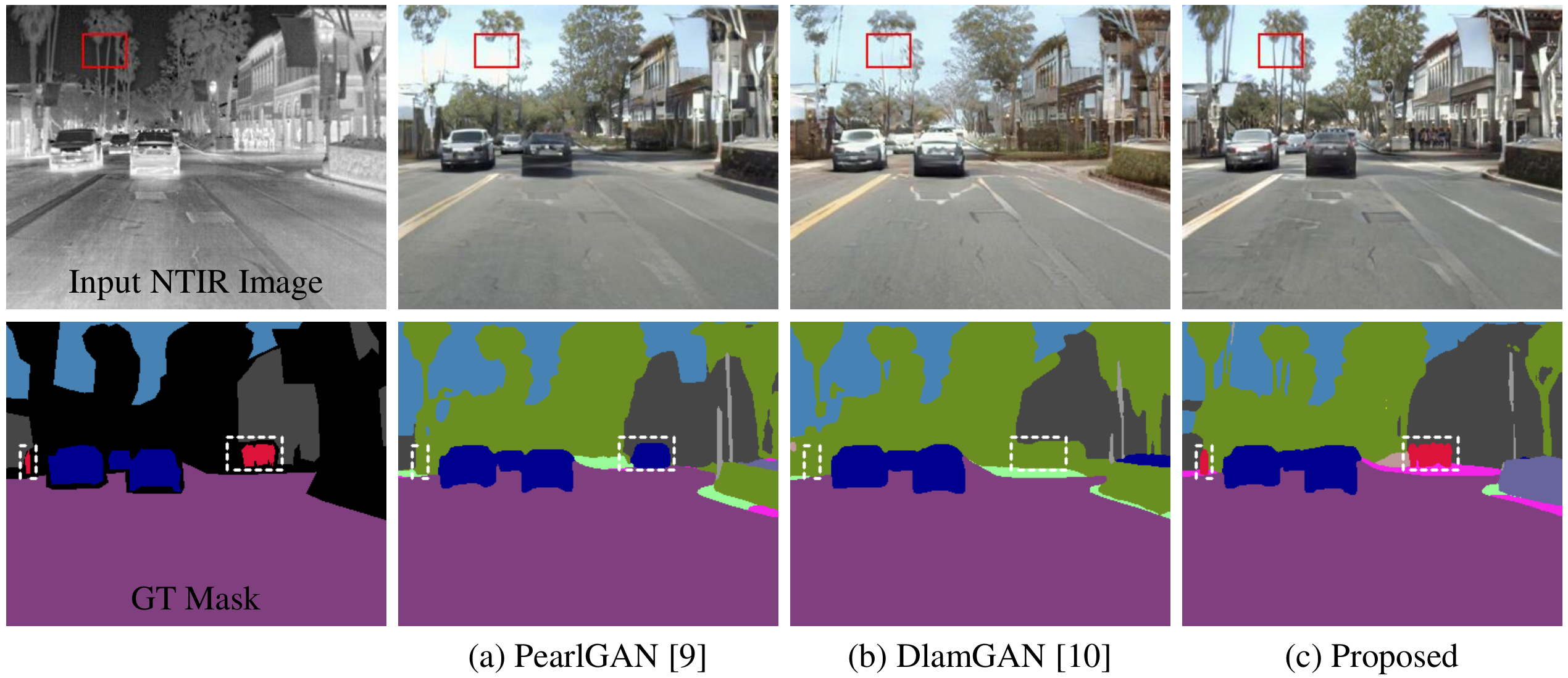}
\caption{Visual comparison of colorization results (the first row) and semantic consistency 
(the second row). The areas in the red boxes and the white dotted boxes deserve attention.}
\label{fig_intro}
\end{figure}

Although encouraging progress has been made in semantically consistent I2I translation, 
three important issues have not been fully considered. First, there has been little research 
on how to improve the texture realism of small sample objects (e.g., pedestrians and traffic 
signs) when there are no available semantic annotations for both domains. Second, how 
to reduce the high confidence noise in pseudo-labels when there is no available semantic 
annotation remains under-explored. Third, the problem of image gradient disappearance in 
local regions during translation is still under-addressed. As shown in Fig. \ref{fig_intro}, the popular 
NTIR2DC methods (e.g., PearlGAN \cite{2022-TITS-Luo} and 
DlamGAN \cite{2021-ICIG-Luo})\footnote{As few available NTIR2DC methods 
exist, only the two methods mentioned are utilized here for comparison.} fail 
to generate plausible pedestrians, as shown in the white dashed boxes. Moreover, the gradients 
of some trunk regions in their colorization results are vanishing, as shown in the red boxes. 
To address the above problems, we propose a Memory-guided cOllaboRative atteNtion GAN 
model (MornGAN).

\begin{figure}[!t]
\centering
\includegraphics[width=3.45in]{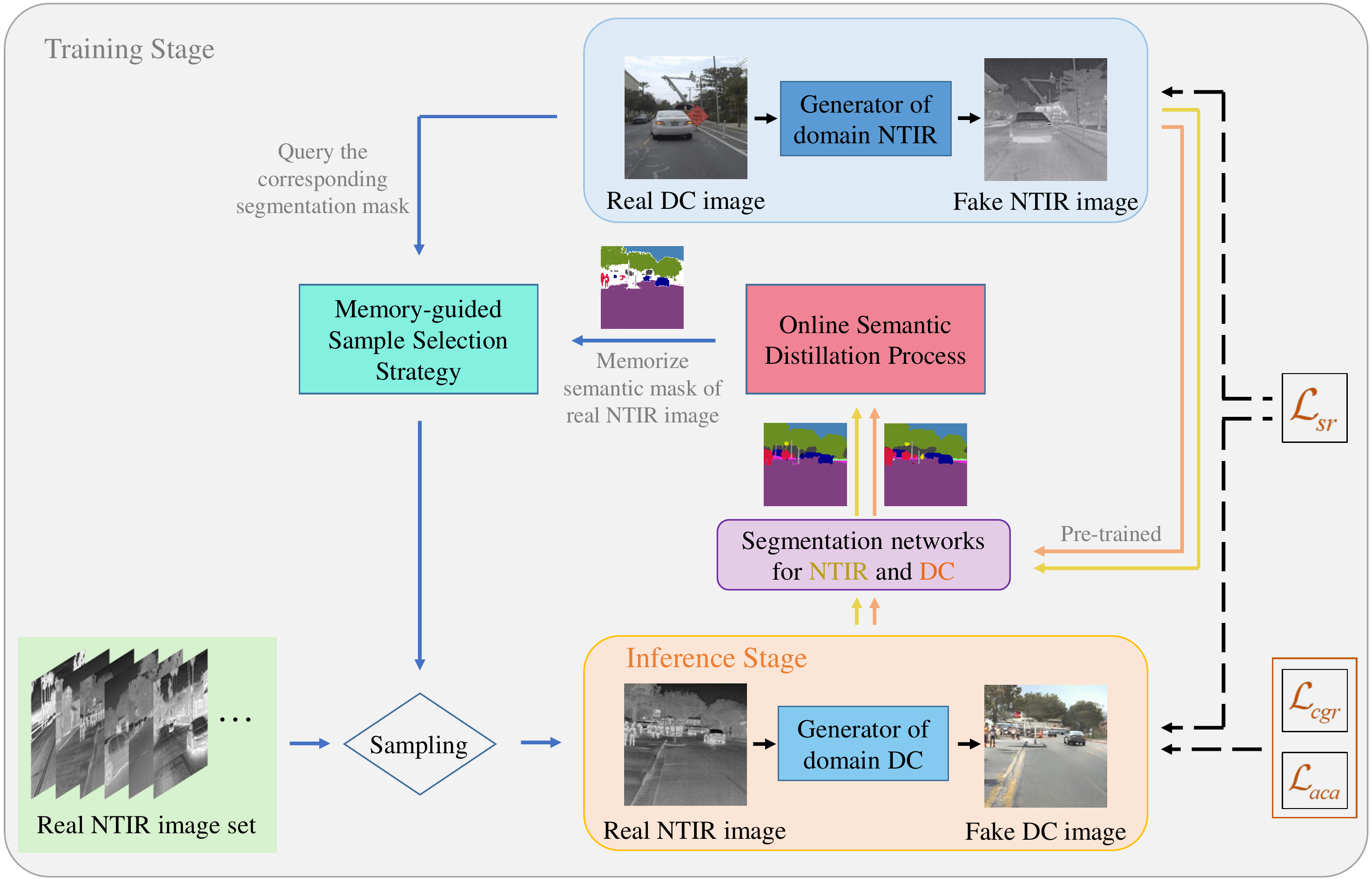}
\caption{Brief illustration of the proposed MornGAN framework. $\mathcal{L}_{aca}$, 
$\mathcal{L}_{cgr}$, and $\mathcal{L}_{sr}$ denote the adaptive collaborative attention loss, 
conditional gradient repair loss, and scale robustness loss, respectively. First, the segmentation 
model obtained by pre-training with real DC and the corresponding fake NTIR image is combined 
with the online semantic distillation process to predict the pseudo-labels of NTIR image, which is 
subsequently stored in the memory unit. Then, the memory-guided sample selection strategy 
associates NTIR images with similar small sample class (e.g., traffic signs in the figure) 
distributions for the input DC image. Finally, the sampled cross-domain image pairs are 
used for learning under the constraints of the corresponding loss functions.}
\label{fig_model_overview}
\end{figure}

We observe that the poor translation performance of small sample objects comes mainly from two 
aspects: the small number of total pixels and the difficulty of learning complex and 
variable texture features. Unlike deep neural networks, humans can efficiently identify a 
common relational system between two contexts and use these commonalities to make further 
inferences, which is called analogical reasoning \cite{2012-Gentner}. Analogical reasoning is 
an important cognitive mechanism that involves retrieving structured knowledge from 
long-term memory and representing the binding of role-fillers in working memory \cite{2013-Oxford-Holyoak}. 
Therefore, inspired by analogical reasoning mechanisms, we design a memory-guided 
collaborative attention approach to improve the translation performance, whose framework is 
shown in Fig. \ref{fig_model_overview}. In this framework, the semantic information of NTIR 
images is first obtained by an online semantic distillation process and then 
memorized online. Subsequently, cross-domain sample pairs containing similar objects are 
associated by a memory-guided sample selection strategy for collaborative learning. Finally, 
adaptive collaborative attention loss is introduced to encourage similarity in the feature 
distributions of objects in the same categories.

To reduce the high-confidence noise in the pseudo-labels of NTIR images, we 
devise an online semantic distillation module that consists of a label mining process 
and a semantic denoising process. The label mining process extracts the high-confidence part 
of the intersection of the segmentation predictions of two domains as 
the coarse labels. Then, the semantic denoising process refines the coarse labels using 
the distributional properties of the original NTIR images. To compensate for edge 
smoothing, we propose a conditional gradient repair 
loss to encourage the preservation of necessary edges. In addition, scale robustness 
loss is introduced to improve the robustness of the model for multi-scale objects.

The main contributions of this study are summarized as follows:
\begin{itemize}
  \item We propose a memory-guided sample selection strategy and an adaptive 
  collaborative attention loss to improve the translation performance of small sample objects, 
  which may provide novel research insights for few-shot domain adaptation and domain generalization.
  \item An online semantic distillation module is devised to mine pseudo-labels for NTIR images, 
  where the semantic denoising process can be generalized to other domains (e.g., visible 
  spectrum) for pseudo-labels refinement.
  \item A conditional gradient repair loss is introduced to reduce edge disappearance during 
  translation, which is important for scene layout preservation.
  \item Extensive experiments on the FLIR \cite{2019-FLIR-FA} and KAIST \cite{2015-CVPR-Hwang} 
  datasets show that the proposed MornGAN\footnote{The source code will be 
  available at \url{https://github.com/FuyaLuo/MornGAN/}.} 
  significantly outperforms other I2I translation methods in terms of semantic preservation 
  and edge consistency, which remarkably improves the object detection accuracy.
\end{itemize}

The rest of this paper is organized as follows. Section 2 summarizes related work about TIR image colorization and I2I 
translation. Section 3 introduces the architecture of the proposed MornGAN. Section 4 presents 
the experiments on the FLIR and KAIST datasets. Section 5 draws the conclusions.

\section{Related Work}
In this section, we briefly review previous work on TIR image colorization, unpaired 
I2I translation, and memory networks.
\subsection{TIR Image Colorization}
TIR image colorization aims to map a single-channel grayscale TIR image to a three-channel RGB image 
based on the image content. With the recent successes of deep neural networks, a large number 
of methods have been proposed to handle TIR image colorization. In general, these methods 
can be classified as supervised or unsupervised methods. Supervised methods usually rely on matched 
cross-domain image pairs to maximize the similarity of the network output to the labels. For 
example, Berg \textit{et al.} \cite{2018-Berg-CVPRW} leveraged separate luminance and 
chrominance loss to optimize the mapping of TIR images to colored visible images. In order 
to increase the naturalness of the results, researchers have made more 
attempts \cite{2020-Bhat-ICCES,2020-Kuang-IPT,2021-ISEE-Le} to colorize TIR images 
based on pixel-level content loss by introducing additional adversarial loss. However, the 
difficulty of collecting pixel-level aligned paired samples limits the practicality of the 
supervised methods for TIR colorization tasks. In contrast, as they do not require paired samples, 
unsupervised methods usually utilize GAN models to make the generated images 
indistinguishable from real RGB images. For example, Nyberg \textit{et al.} \cite{2018-Nyberg-ECCV} 
exploited the CycleGAN \cite{2017-CVPR-Zhu} model to realize unpaired infrared-visible image 
translation. PearlGAN \cite{2022-TITS-Luo} was proposed to reduce semantic encoding entanglement 
and geometric distortion in the NTIR2DC task. DlamGAN \cite{2021-ICIG-Luo} was designed with a 
dynamic label mining module to predict the semantic masks of NTIR images to encourage semantically 
consistent colorization. Despite the impressive progress, few efforts have been made to 
improve the colorization performance of small sample objects.

\subsection{Unpaired I2I Translation}
The purpose of unpaired I2I translation is to learn the mapping functions between 
different image domains using unpaired samples. Driven by the cycle consistency loss in 
CycleGAN \cite{2017-CVPR-Zhu}, the unpaired I2I translation task has gained 
considerable attention in the computer vision community \cite{2017-NIPS-Liu,2020-ECCV-Zheng,2020-CVPR-Chen}. 
For example, MUNIT \cite{2018-ECCV-Huang} and DRIT++ \cite{2020-IJCV-Lee} were proposed to 
improve the diversity of synthesized images by learning a disentangled style representation 
and content representation. Anoosheh \textit{et al.} \cite{2019-ICRA-Anoosheh} utilized multiple 
discriminators to improve the generation performance of night-to-day image translation. To 
reduce content distortion during translation, many 
researchers \cite{2018-ECCV-Huang-Auggan,2019-WACV-Cherian,2020-WACV-Pizzati} have 
introduced semantic consistency loss using the available semantic annotations. When no semantic 
annotation is available for both domains, DlamGAN \cite{2021-ICIG-Luo} first predicts the 
pseudo-labels of one domain using domain adaptation, and then introduces a dynamic label mining 
module to obtain the pseudo-labels of the other domain. Although semantic consistency loss can 
significantly reduce the semantic distortion during translation, the edges within or between 
classes of the background category are usually smoothed or disappear to enhance the realism 
of the image patches. However, 
this image texture vanishing problem is usually underappreciated. 
Moreover, how to reduce the high confidence noise in pseudo-labels using domain knowledge is 
still under-explored.

\subsection{Neural Networks with External Memory}
A memory network \cite{2015-ICLR-Weston,2015-NeurIPS-Sukhbaatar} is a learnable neural network 
module that allows writing information to external memory and reading relevant content from 
memory slots. Due to their storage of long-term information and explicit memory manipulation, 
memory networks have been widely adopted in solving various computer vision problems such as few-shot 
learning \cite{2021-ICCV-Xie,2021-ICCV-Wu}, semi-supervised learning \cite{2021-ICCV-Alonso}, 
domain adaptation \cite{2021-CVPR-VS}, and domain generalization \cite{2021-ICCV-Chen}. For 
example, Xie \textit{et al.} \cite{2021-ICCV-Xie} proposed a recurrent memory network to directly learn 
to recursively read information from the support set features at all resolutions and capture 
features across resolutions to achieve more accurate few-shot semantic segmentation. 
Alonso \textit{et al.} \cite{2021-ICCV-Alonso} used a memory bank to store and update the 
category-level features of labeled data, and constrain the category-level features of unlabeled 
data to be consistent with the memorized feature to achieve semi-supervised semantic segmentation. 
Furthermore, to improve the transfer performance of object features, VS \textit{et al.} \cite{2021-CVPR-VS} 
exploited memory-guided attention maps to route target domain features into the corresponding 
category discriminators to ensure the domain alignment of category features. 
Jeong \textit{et al.} \cite{2021-CVPR-Jeong} proposed a class-aware memory network to explicitly 
record category-level style differences for instance-level image translation using bounding-box 
annotation. Unlike existing methods, the proposed method does not require manual annotation of 
training data, and the introduction of memory units does not increase the computational cost of 
the inference stage.

\section{Proposed Method}

\begin{figure*}[!t]
\centering
\includegraphics[width=1\textwidth]{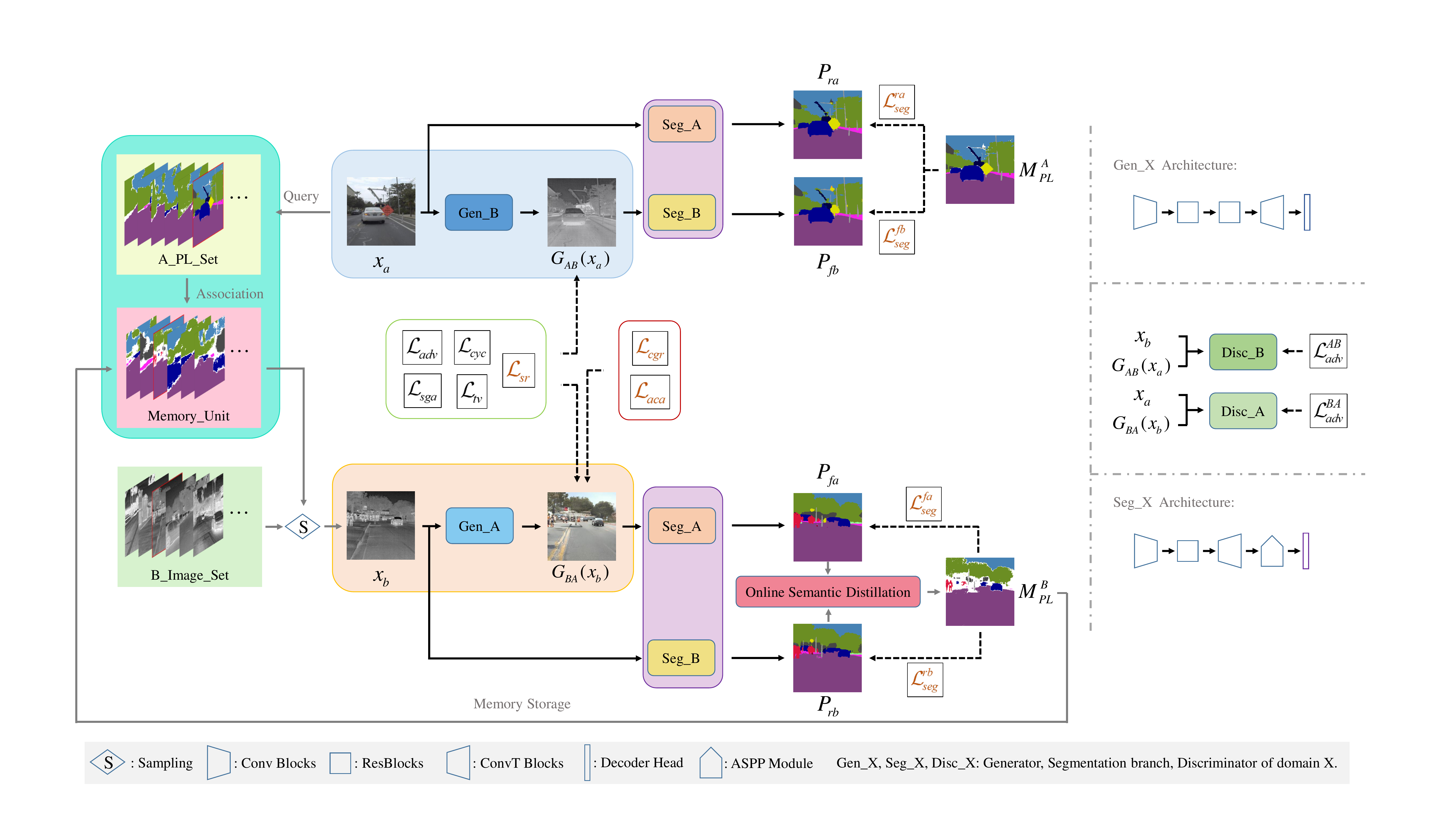}
\caption{The overall architecture of the proposed method. $x_a$ is a random image in daytime 
visible domain A, while $x_b$ is an image in the NTIR domain B sampled according to a memory-guided 
sample selection strategy. The memory unit stores the pseudo-labels $M_{PL}^{B}$ of the NTIR image 
predicted by the online semantic distillation module. Then, the sampled cross-domain image 
pairs are jointly learned under the constraints of the corresponding loss functions.}
\label{fig_model}
\end{figure*} 

In this section, we first present the overview of MornGAN. Subsequently, we 
briefly explain our baseline model. Then, the details of online semantic distillation 
module are described. Next, we explicate the memory-guided collaborative attention mechanism, 
including the memory-guided sample selection strategy and adaptive collaborative 
attention (ACA) loss. Afterward, the conditional gradient repair (CGR) loss constraining edge 
consistency during translation is explained. Then, the scale robustness (SR) loss 
responsible for boosting the robustness of the model to object scale changes is specified. 
Finally, we illustrate the total loss of MornGAN.

In the rest of the paper, domain A and domain B denote the DC image set and NTIR image set, 
respectively. Taking the translation from domain A to domain B as an example, we denote the input 
image pair of domain A and B as $\left \{ x_{a}, x_{b} \right \} $, the 
generator $G_{AB}$ contains an encoder of domain A and a decoder of domain B, and the 
discriminator $D_B$ aims to distinguish the real image $x_b$ from the translated 
image $G_{AB} \left ( x_{a}  \right ) $. Similarly, the inverse mapping includes 
generator $G_{BA}$ and discriminator $D_A$.

\subsection{Model's Overview and Problem Formulation}

The overall framework is shown in Fig. \ref{fig_model}. We first improve our previously developed 
ToDayGAN-TIR \cite{2022-TITS-Luo} model to accommodate subsequent semantic consistency 
requirements, which is called ToDayGAN-NTIR. Then, the ToDayGAN-NTIR model, containing a pair of 
generators and discriminators, is used as a baseline model with a total objective function 
consisting of adversarial loss $\mathcal{L}_{adv}$, cycle-consistency loss $\mathcal{L}_{cyc}$, 
total variance loss $\mathcal{L}_{tv}$, and structured gradient alignment (SGA) 
loss $\mathcal{L}_{sga}$. Subsequently, we introduce two segmentation networks, $S_{A}$ 
and $S_{B}$, to predict the segmentation masks of the images in both domains. 
The segmented pseudo-labels of the two domains, $M_{PL}^{A}$ and $M_{PL}^{B}$, 
are obtained by the existing semantic segmentation model \cite{2019-detectron2-Wu,2020-Arxiv-Tao} 
of DC images and the proposed online semantic distillation module, respectively. 
The pseudo-labels of NTIR images are subsequently stored in the memory unit. 
When the stored quantity meets the condition, the memory-guided sample selection 
strategy is triggered, and then similar NTIR images are recalled for the 
input DC images for collaborative learning. Finally, ACA loss $\mathcal{L}_{aca}$ constrains 
the inter-domain similarity of 
features of small sample classes. In addition, segmentation losses of synthesized images, 
denoted as $\mathcal{L}_{seg}^{fa}$ and $\mathcal{L}_{seg}^{fb}$, are used to encourage 
semantically invariant image translation. To reduce edge degradation during translation, 
CGR loss $\mathcal{L}_{cgr}$ is introduced. 
The SR loss $\mathcal{L}_{sr}$ aims to improve the insensitivity of the model 
to the object scale.

\subsection{Baseline Model}
\subsubsection{Revisiting ToDayGAN-TIR Model}
The ToDayGAN \cite{2019-ICRA-Anoosheh} model introduces three discriminators to improve the 
translation performance of visible images from nighttime to daytime. 
ToDayGAN-TIR \cite{2022-TITS-Luo} adapts 
ToDayGAN to improve the performance of NTIR image colorization. To avoid dot 
artifacts, ToDayGAN-TIR replaces the last two instance normalization layers of the 
decoder with group normalization \cite{2018-ECCV-Wu} layers and combines the total 
variance \cite{1992-PD-Rudin} loss $\mathcal{L}_{tv}$ to reduce the noise of the translation results. To improve 
the representation ability of cycle-consistency loss $\mathcal{L}_{cyc}$ for NTIR images, it introduces 
SSIM \cite{2004-TIP-Wang} loss based on the original L1-norm loss. In addition, it introduces 
the spectral normalization \cite{2018-Arxiv-Miyato} layer into the discriminator to make the 
model training more stable. Similar to ToDayGAN, it chooses relativistic 
loss \cite{2018-Arxiv-Jolicoeur}, adapted to least-squares GAN loss, as adversarial 
loss $\mathcal{L}_{adv}$.

\subsubsection{Generating High-quality Fake NTIR Images}
To predict the semantic mask of NTIR images without available annotation, which is a necessary 
component for semantic consistency loss, we need high-quality fake NTIR images and 
corresponding DC image pseudo-labels to train the segmentation network for NTIR domain. 
Therefore, we first introduce SGA \cite{2022-TITS-Luo} loss $\mathcal{L}_{sga}^{ori}$ to 
reduce the edge distortion during translation based on the ToDayGAN-TIR model. SGA loss 
encourages the 
ratio of the gradient of the synthesized image at the edge of the original image to the maximum 
gradient to be greater than a given threshold. In addition, to further obtain high-quality 
fake NTIR images, we introduce two regularization terms in the SGA loss of domain A, the 
monochromatic regularization term and the temperature regularization term. As the output fake NTIR image 
is three-channel data, the monochromaticity regularization term aims to encourage the values of the 
three channels to be the same. For the unnatural situation where the mean value of the 
pedestrian area in the fake NTIR image is extremely small, the temperature regularization term is 
responsible for encouraging the minimum value of the pedestrian area to be no less 
than the mean value of the road area. This regularization term is inspired by the observation that 
the body temperature of pedestrians is usually higher than the mean temperature of the road area 
during nighttime conditions.

Concretely, similar to \cite{2019-TVT-Ma}, we first define the channel maximum operation as a 
mapping $cmax:\mathbb{R}^{C\times H\times W}  \to \mathbb{R}^{H\times W}$. Then, given the 
input $I\in \mathbb{R}^{C\times H\times W}$, the output $O\in \mathbb{R}^{H\times W}$ of the 
channel maximum operation at position $\left ( i,j \right ) $ can be expressed as
\begin{equation}
  \label{C_max}
  O_{i,j} =\max_{c\in \left \{ 1,2,\dots ,C \right \} } I_{c,i,j}.  
\end{equation}
Next, the monochromatic regularization term can be denoted as
\begin{equation}
  \label{T_mr}
  T_{mr} =\max \left ( cmax\left(G_{AB}\left(x_{a} \right) \right) -cmin\left(G_{AB}\left(x_{a} \right)\right)\right).  
\end{equation}
For the temperature regularization term, given the mean value of the road region of the fake NTIR image, denoted 
as $\bar{v}_{road}^{fb}$, and the minimum value of the pedestrian region of the fake NTIR image, denoted 
as $\tilde{v}_{ped}^{fb}$, the temperature regularization term can be represented as
\begin{equation}
  \label{T_tr}
  T_{tr} =\max \left(\frac{\bar{v}_{road}^{fb}-\tilde{v}_{ped}^{fb}}{\bar{v}_{road}^{fb}+ \varepsilon} ,0 \right), 
\end{equation}
where the denominator is intended to normalize the output in the 
range $\left [ 0,1 \right ) $, and $\varepsilon$ is a small value to avoid dividing by zero. 
Ultimately, the improved SGA loss can be expressed as
\begin{equation}
  \label{L_sga}
  \mathcal{L}_{sga} =\mathcal{L}_{sga}^{ori}+ T_{mr} + T_{tr}.  
\end{equation}
Combining the above two adjustments, we obtain the variant model ToDayGAN-NTIR, which 
serves as the baseline for MornGAN.

\subsection{Pseudo-label Inference and Segmentation Loss}
The above ToDayGAN-NTIR model still does not solve the problem of content distortion in the 
translation process. Therefore, similar to \cite{2018-ECCV-Huang-Auggan,2019-WACV-Cherian}, we 
introduce auxiliary segmentation networks (i.e, $S_{A}$ and $S_{B}$) and segmentation masks to 
encourage semantically consistent I2I translation. Due to the lack of manual annotation of both 
domains, similar to DlamGAN \cite{2021-ICIG-Luo}, we first obtain the pseudo-labels of 
domain A using the existing DC image segmentation model, and then design a label 
mining module to predict the pseudo-labels of NTIR images online.

Unlike DlamGAN, our proposed approach uses an online semantic distillation module that not only 
exploits different thresholds to balance the distribution bias among categories, but also 
utilizes the variation in temperature distribution among categories to reduce high-confidence 
noise.

\subsubsection{Semantic Denoising}
\begin{figure}[!t]
\centering
\includegraphics[width=3.45in]{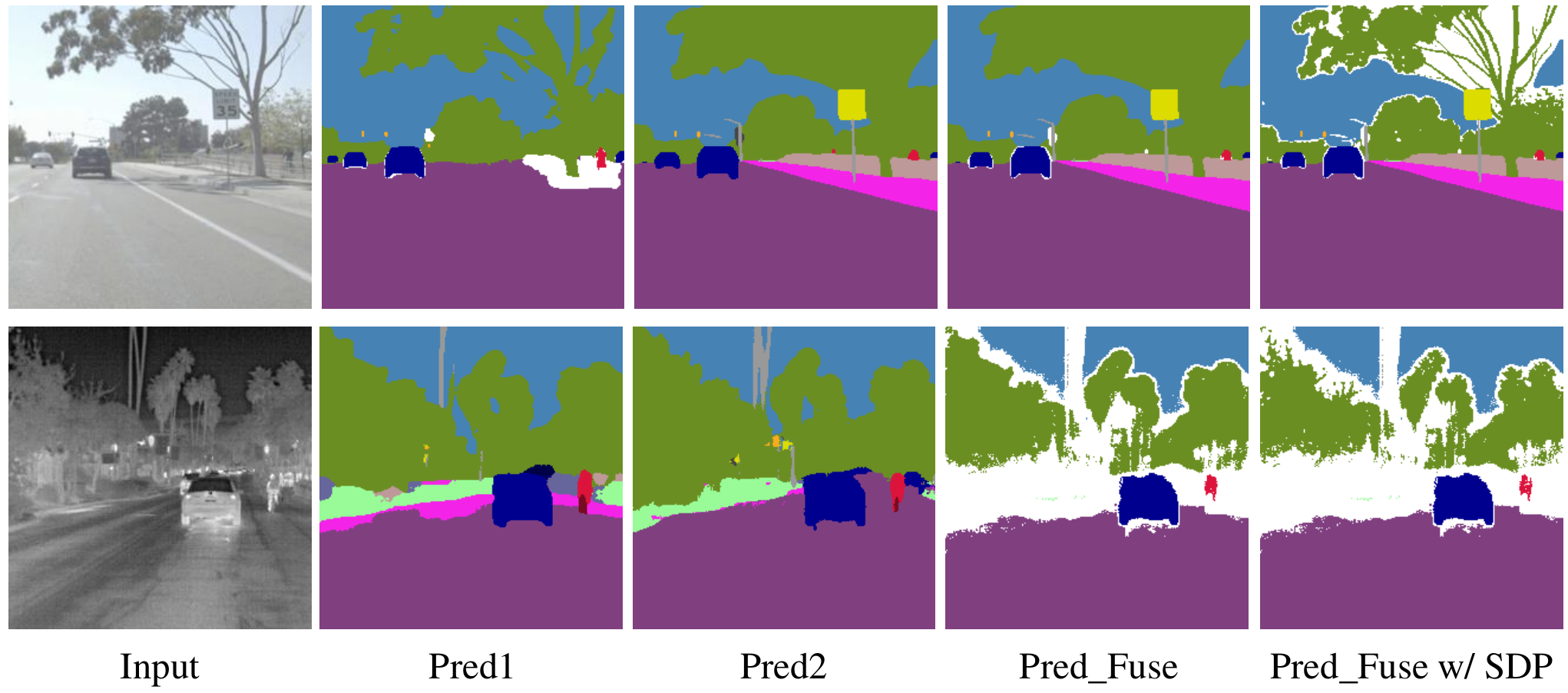}
\caption{Examples of pseudo-label inference and semantic denoising on two domains. 
Pred1 and Pred2 in the first row are derived from the predictions of 
Detectron2 \cite{2019-detectron2-Wu} and HMSANet \cite{2020-Arxiv-Tao}, respectively. 
Pred1 and Pred2 in the second row are derived from the predictions of the 
corresponding fake DC images and the input NTIR images, respectively. Then, the results in 
the fourth column are obtained by integrating Pred1 and Pred2 based on the domain-specific 
fusion rules. The last column lists the semantic denoising results of the fourth column.}
\label{fig_sdp}
\end{figure}

To obtain segmentation pseudo-labels for both domains, we first introduce a novel semantic 
denoising process that utilizes the class-specific low-rank properties of the original image 
to remove noisy labels that deviate from the distribution. For the specific category $y_{1}$ in 
the dataset, given the coarse pseudo-labels $M_{all} \in \mathbb{R} ^{H\times W}$, the input 
image $I_{x} \in \mathbb{R} ^{C\times H\times W}$, and the set 
$Z_{y_{1}} =\left \{ y_{12} ,y_{13},\cdots ,y_{1m} \right \} $ of confusion categories 
of $y_{1}$, we can obtain a more trustworthy binary 
mask $\hat{M}_{y_{1}} \in \mathbb{R}^{H\times W}$ through the semantic denoising process. 
Specifically, for a given category $y_{k}$, we first compute its binary 
mask $M_{y_{k}} \in \mathbb{R}^{H\times W}$ and average pixel 
feature $f_{y_{k}} \in \mathbb{R} ^{C\times 1}$. Subsequently, 
we define the matrix of feature distances from pixels belonging to category $y_{1}$ to category 
$y_{k}$ as $Q_{y_{1}y_{k}}$, and the value of $Q_{y_{1}y_{k}}$ at position $u$ can be formulated 
as
\begin{equation}
  \label{Q_y1yk}
  \left(Q_{y_{1}y_{k}}\right)_{u} =\left(M_{y_{1}}\right)_{u}\times \sum \left(\left(M_{y_{1}} \odot I_{x}\right)_{u} -f_{y_{k}}\right)^{2},  
\end{equation}
where $\odot$ denotes element-wise multiplication with channel-wise broadcasting, and the 
first term of the multiplication aims to ignore the feature distances of the 
locations that do not belong to category $y_{1}$, while the second term is used to calculate 
the Euclidean distance between features. Ultimately, the value of $\hat{M}_{y_{1}}$ at position 
$u$ can be given by
\begin{equation}
  \label{M_y1new}
  \begin{split}
    \left(\hat{M}_{y_{1}}\right)_{u} =&\max \left(\left(\left(M_{y_{1}}\right)_{u} - \left(\mathbb{I}\left\{\left(Q_{y_{1}y_{1}} -Q_{y_{1}y_{12}} \right)_{u} > 0 \right\}+ \right. \right. \right. \\
    & \left. \left. \left. \cdots +  \mathbb{I}\left\{\left(Q_{y_{1}y_{1}} -Q_{y_{1}y_{1m}} \right)_{u} > 0 \right\}\right)\right) ,0\right),
  \end{split} 
\end{equation}
where $\mathbb{I}\left \{\cdot\right\}$ is the indicator function (i.e., output 1 when the 
condition is met, and otherwise 0), and the equation in the indicator function determines whether 
the label at that location is 
noise by comparing the magnitude of the intra-class distance with that of the inter-class 
distance. In sum, the above semantic denoising process can be expressed using the 
mapping $\mathcal{SDP}\left( \cdot \right)$ as
\begin{equation}
  \label{sdp}
  \hat{M}_{y_{1}}=\mathcal{SDP}\left(M_{all} ,I_{x},Z_{y_{1} }\right).   
\end{equation}
If there are multiple categories to be denoised, each category is updated with its own 
average feature after denoising, which enables a more 
reasonable estimation of the inter-class distance afterward. Two examples of SDP are 
shown in Fig. \ref{fig_sdp}, and the fifth column shows the result of SDP on the fourth column.

\subsubsection{Pseudo-label Inference of Visible Domain}
As there is no segmentation label for DC images and the domain-adapted semantic segmentation 
methods still suffer from high-confidence noise, we directly utilize the existing segmentation 
models trained on different datasets to jointly predict the pseudo-labels of DC images. 
Concretely, we choose Detectron2 \cite{2019-detectron2-Wu}, a panoptic segmentation model 
trained on the MS COCO dataset \cite{2014-ECCV-Lin}, and HMSANet \cite{2020-Arxiv-Tao}, a 
semantic segmentation model trained on the Cityscape dataset \cite{2016-CVPR-Cordts}. Due to 
the difference in training data, the Detectron2 model is good at object contour segmentation 
but is weaker than the HMSANet model for segmentation of traffic scene categories, as shown in 
the first row in Fig. \ref{fig_sdp}. Therefore, we take the intersection of the predictions of HMSANet 
and Detectron2 
as the pseudo-labels of the pedestrian, car, and building categories, and the predictions of 
HMSANet for the other categories. The pseudo-labels obtained by fusion are noted as $M_{F}^{A}$.

However, we observe that the segmentation results after upsampling usually show shifting 
of edges and inflation of semantic regions due to the pooling 
and strided convolution operations in deep neural networks. In addition, there are significant 
low-rank properties (e.g., color homogeneity) within some background classes (i.e., tree, sky, 
and pole) in the visible domain with large inter-class differences. Accordingly, we exploit the 
distributional properties of these categories to reduce the noise in the pseudo-labels. 
Specifically, assuming that the categories of tree, pole, and sky are denoted 
as $t$, $o$ and $s$, respectively, the 
pseudo-labels of these three categories are refined sequentially using the semantic denoising 
process mentioned above; that is, $\mathcal{SDP}\left(M_{F}^{A},x_{a},Z_{t}\right)$, 
$\mathcal{SDP}\left(M_{F}^{A},x_{a} ,Z_{o}\right)$ and $\mathcal{SDP}\left(M_{F}^{A},x_{a} ,Z_{s}\right)$. 
Because the semantic edges of trees and poles usually expand into the sky region in the predicted 
segmentation masks, the set of confusion categories $Z_{t}$, $Z_{o}$, and $Z_{s}$ are 
$\left\{ s \right\}$, $\left\{ s \right\}$, and $\left\{ t, o \right\}$, respectively. 
After denoising, we can obtain the pseudo-labels $M_{PL}^{A}$ of the DC image. An 
example image after denoising is shown in the fifth column of the first row in Fig. 4, and the 
final pseudo-labels can significantly reduce the predicted noise.

\subsubsection{Online Semantic Distillation Module}
With the obtained pseudo-labels of DC images, we can train the segmentation network $S_{A}$ 
with input $x_{a}$, and $S_{B}$ with input $G_{AB}\left(x_{a}\right)$. After that, 
we devise an online semantic distillation module to 
jointly predict the pseudo-labels of NTIR images by using $S_{A}$ and $S_{B}$. The online semantic 
distillation module consists of a label mining process and a semantic denoising process, where 
the former mines the high-confidence part of the network prediction intersection as the 
coarse pseudo-labels, and the latter denoises the pseudo-labels based on the category 
distribution properties. Concretely, we first define the probability tensor of outputs 
$S_{B}\left( x_{b} \right)$ and $S_{A}\left(G_{BA} \left(x_{b}\right)\right) $ as 
$V^{rb}\in \mathbb{R} ^{N_{c}\times H\times W}$ and 
$V^{fa}\in \mathbb{R} ^{N_{c}\times H\times W}$, respectively, whose values at 
the $c_{th}$ channel and at position $u$ denote the probability that the position belongs 
to category $c$, denoted as $V_{c,u}^{rb}$ and $V_{c,u}^{fa}$. Moreover, $N_{c}$ denotes the number 
of categories. Considering the difference in the number of categories and the fact that GAN 
models usually translate object regions in IR images into background regions to enhance the 
realism of synthesized images \cite{2022-TITS-Luo}, label mining using the same threshold for 
all categories is suboptimal. Therefore, we design two thresholds, $\theta_{fg}$ and 
$\theta_{bg}$, to extract the pseudo-labels of the foreground category $\mathcal{C}_{fg}$ and 
background category $\mathcal{C}_{bg}$, respectively. Then, the category of pseudo-labels at 
location $u$ obtained by the label mining process, denoted as $\left(M_{LM}^{B} \right)_{u}$, 
can be represented as

\begin{equation}
  \label{M_LB}
  \begin{split}
    \left(M_{LM}^{B} \right)_{u} =
  \begin{cases}
    c{}', & \mbox{if $V_{c{}',u}^{rb}\ge \eta \left(c{}'\right),V_{c{}',u}^{fa}\ge \eta \left(c{}'\right).$} \\
    unlabeled, & \mbox{otherwise.}
  \end{cases}
  \end{split} 
\end{equation}
And $\eta \left(\cdot \right)$ is a category-dependent piecewise function:
\begin{equation}
  \label{th_LB}
  \begin{split}
    \eta\left( c\right) =
  \begin{cases}
    \theta_{fg}, & \mbox{if $c\in \mathcal{C}_{fg}.$} \\
    \theta_{bg}, & \mbox{otherwise.}
  \end{cases}
  \end{split} 
\end{equation}

Due to the presence of high confidence noise in the label mining results, we exploit the 
semantic denoising process to suppress the noisy labels similar to how we handle pseudo-labels 
of DC images. 
Considering that the body temperature of pedestrians is usually significantly higher than 
some of the background categories (e.g., trees) in the nighttime environment, we add denoising 
for the pedestrian region to the three background categories (i.e., trees, poles, and sky) 
mentioned in the previous subsection. Specifically, we first abbreviate the pedestrian category 
as $p$. Then, we use Eq. (\ref{sdp}) to denoise the four categories of sky, pole, pedestrian 
and tree in turn, i.e., $\mathcal{SDP}\left(M_{LM}^{B},x_{b},Z_{s}\right)$, 
$\mathcal{SDP}\left(M_{LM}^{B},x_{b},Z_{o}\right)$, $\mathcal{SDP}\left(M_{LM}^{B},x_{b},Z_{p}\right)$, 
$\mathcal{SDP}\left(M_{LM}^{B},x_{b},Z_{t}\right)$. Considering the spatial distribution of the 
categories, the set of confusion categories $Z_{s}$, $Z_{o}$, $Z_{p}$, and $Z_{t}$ is set to 
$\left\{t, o\right\}$, $\left\{s\right\}$, $\left\{t\right\}$, and $\left\{s, p\right\}$, 
respectively. After denoising, we can obtain the pseudo-labels of the NTIR image, denoted 
as $M_{PL}^{B}$. An example diagram of the online semantic distillation module is shown in 
the second row in Fig. \ref{fig_sdp}.

\subsubsection{Segmentation Loss}
Pseudo-labels of both domains are used not only for supervising the training of the 
segmentation network, but also for encouraging semantically invariant image translation. 
Both stages require segmentation loss as the optimization objective. Due to the uneven 
distribution among categories, we utilize a modified pixel-level cross-entropy 
loss \cite{2019-IJCAI-Zheng} as the segmentation loss for the three 
branches (i.e., $\mathcal{L}_{seg}^{fb}$, $\mathcal{L}_{seg}^{rb}$, and $\mathcal{L}_{seg}^{fa}$), 
which assigns larger weights for smaller sample categories. Thanks to the sharp edges in the DC 
images and the relatively complete semantic edges in the pseudo-labels, we can improve the edge 
segmentation performance of model $S_{A}$ by introducing boundary loss. Referring 
to \cite{2020-CVPR-Zhen}, the absolute value of the difference between the original image and 
the result after average pooling is taken as the spatial gradient. Then, we can calculate 
the difference in the spatial gradient between the predicted semantic probability tensor and the 
label, whose absolute value is taken as the boundary loss. Thus, the 
loss $\mathcal{L}_{seg}^{ra}$ consists of a modified pixel-level cross-entropy 
loss \cite{2019-IJCAI-Zheng} and a boundary loss \cite{2020-CVPR-Zhen}, where the weight of 
the boundary loss is 
empirically set to 0.5. Ultimately, the complete segmentation loss can be expressed as
\begin{equation}
  \label{L_seg}
  \mathcal{L}_{seg}^{all} =\lambda_{ra} \mathcal{L}_{seg}^{ra}+ \lambda_{fb} \mathcal{L}_{seg}^{fb}+\lambda_{rb}\mathcal{L}_{seg}^{rb}+\lambda_{fa}\mathcal{L}_{seg}^{fa}, 
\end{equation}
where $\lambda_{ra}$, $\lambda_{fb}$, $\lambda_{rb}$, and $\lambda_{fa}$ are 
binary (i.e., either 0 or 1) loss weights used to switch learning stages.

\subsection{Memory-guided Collaborative Attention}
\label{subsec_MCA}
Although the online semantic distillation module and segmentation loss can help reduce the 
semantic distortion during translation, how to improve the translation performance of small 
sample categories remains to be explored. We observe that regions of small-sample categories 
in DC images have relatively complete semantic masks, which are usually fragmented 
for NTIR images. However, humans are usually good at combining memory and similarities between 
two situations to solve such small sample inference problems \cite{2012-Gentner,2013-Oxford-Holyoak}, 
which is called the analogical reasoning mechanism. Inspired by this insight, we design 
a memory-guided collaborative attention mechanism to improve the colorization performance of 
small sample categories $\mathcal{C}_{ss}$, which consists of a memory-guided sample 
selection strategy and an adaptive collaborative attention loss.
\subsubsection{Memory-guided Sample Selection Strategy}
The memory-guided sample selection strategy aims to select cross-domain sample pairs 
containing objects of the same categories for collaborative learning. Due to the lack 
of semantic labels of NTIR images, 
online memorization of semantic masks of NTIR images is necessary. Therefore, as shown in 
Fig. \ref{fig_model}, after the weights of the segmentation network are fixed, we leverage the online 
semantic distillation module to infer the pseudo-label $M_{PL}^{B}$ of the NTIR image, which 
is subsequently stored in the memory unit. After the semantic masks of all NTIR images are 
stored, the sample selection strategy works to select the appropriate NTIR images for 
a given DC image based on the distribution similarity. 

As the goal is to improve the colorization 
performance of small sample categories, the focus is on the similarity of the distribution of 
cross-domain sample pairs in small sample categories. Concretely, 
given the semantic mask $M_{PL}^{A}$ of a DC image, we first calculate the percentage of regions 
corresponding to each small sample category relative to the full image, and then subtract 
the mean value to obtain the vector $f_{d}^{A}$ as the semantic distribution feature. 
Similarly, we can obtain the semantic distribution feature $f_{d}^{B}$ of any NTIR image. 
Further, the similarity of the distribution between $f_{d}^{A}$ and $f_{d}^{B}$, 
denoted as $d_{AB}$, can be expressed using the folldue cosine similarity:
\begin{equation}
  \label{d_AB}
  d_{AB} =\frac{f_{d}^{A}\cdot f_{d}^{B}}{\left \| f_{d}^{A} \right \|_{2} \cdot \left \| f_{d}^{B} \right \|_{2}}, 
\end{equation}
where $\left\| \cdot \right\| _{2}$ represents the L2 norm.

Although we can use similarity to find the NTIR image with the most similar (i.e., top-1 
selection) distribution for the given DC image, the incompleteness of $M_{PL}^{B}$ may lead to 
frequent selection of a particular NTIR image, which can cause overfitting of the model. To 
avoid this problem, the top-1 selection strategy is 
relaxed to random sampling from the top-k candidates, which means that one of the $k$ most 
similar NTIR images is randomly selected for collaborative learning with the given DC image. 
Unlike the popular learning method of random sample pairs in GAN 
models \cite{2014-NIPS-Goodfellow}, the proposed collaborative learning approach can pave the 
way for subsequent category-aware cross-domain constraints.

\subsubsection{Adaptive Collaborative Attention (ACA) Loss}
With the selection of cross-domain sample pairs containing small sample categories, an ACA loss is 
designed to further encourage the similarity of feature distributions within classes. Due to the 
complexity of the constituent parts of object categories, characterizing the texture of 
objects with a single mean feature is sub-optimal. Accordingly, we first use 
Kmeans \cite{1967-BSM-MacQueen,1982-TIT-Lloyd} to extract the features of the components of 
the small sample categories of the real DC image. Then, the inner product between 
each component feature and the original feature is used to characterize the response 
map or co-attentive map of that component feature. Ultimately, the distance between the 
corresponding response maps of the fake and DC images is used to 
measure the similarity of the feature distributions. Specifically, we first denote the 
features of the real DC image and the fake DC image after the encoder as 
$F^{ra} \in \mathbb{R} ^{C\times h\times w}$ and $F^{fa} \in \mathbb{R} ^{C\times h\times w}$, 
respectively, and their corresponding segmentation pseudo-labels as 
$M^{A} \in \mathbb{R}^{h\times w}$ and $M^{B} \in \mathbb{R} ^{h\times w}$, respectively. 
Given any category $c_{k}$ in the small sample category set $\mathcal{C}_{ss}$, we can obtain 
the binary masks $M_{c_{k}}^{A}$ and $M_{c_{k}}^{B}$ of this category in two domains. Then, 
let the 
sum of non-zero elements in $M_{c_{k}}^{A}$ be $N_{c_{k}}^{A}$, the feature matrix 
$F_{c_{k}}^{ra} \in \mathbb{R} ^{C\times N_{c_{k}}^{A}}$ corresponding to the category $c_{k}$ 
in the real DC image can be denoted as
\begin{equation}
  \label{F_ck}
  F_{c_{k}}^{ra}=\mathcal{E} \left (M_{c_{k}}^{A} \odot F^{ra} \right), 
\end{equation}
where $\mathcal{E} \left(\cdot \right)$ denotes the operation of first transforming the input 
into a matrix of $C$ rows $h \times w$ columns and then extracting the columns in which the 
sum of absolute values is non-zero. Subsequently, we utilize Kmeans to obtain the clustering 
feature matrix $U_{c_{k}}^{ra} \in \mathbb{R}^{N_{u}\times C}$, where each row denotes the 
features of the cluster centroids, and $N_{u}$ denotes the number of clusters. Afterward, we 
can obtain the response map $Y_{c_{k}}^{ra} \in \mathbb{R} ^{N_{u}\times N_{c_{k}}^{A}}$ of 
all centroid features, which can be expressed using the cosine similarity between 
features as follows:
\begin{equation}
  \label{Y_ck}
  Y_{c_{k}}^{ra} =\frac{U_{c_{k}}^{ra}\times F_{c_{k}}^{ra}}{\left \| U_{c_{k}}^{ra} \right \|_{2} \cdot \left \| F_{c_{k}}^{ra} \right \|_{2}}. 
\end{equation}
Then, we reshape $Y_{c_{k}}^{ra}$ into 
$\tilde{Y}_{c_{k}}^{ra}\in \mathbb{R} ^{N_{u}\times N_{c_{k}}^{A}\times 1}$ and 
$\hat{Y}_{c_{k}}^{ra}\in \mathbb{R} ^{ N_{c_{k}}^{A}\times N_{u}\times1}$. Further, the mean 
value of the maximum response of each location for all centroid features can be 
formulated as
\begin{equation}
  \label{u_ck}
  \mu_{c_{k}}^{ra}=\frac{1}{N_{c_{k}}^{A}} \sum cmax\left ( \tilde{Y}_{c_{k}}^{ra} \right). 
\end{equation}
Correspondingly, the mean value of the maximum response of each cluster for the features at 
all locations can be expressed as
\begin{equation}
  \label{t_ck}
  \tau_{c_{k}}^{ra}=\frac{1}{N_{u}} \sum cmax\left ( \hat{Y}_{c_{k}}^{ra} \right). 
\end{equation}
Similarly, we can compute $\mu_{c_{k}}^{fa}$ and $\tau_{c_{k}}^{fa}$ using $F^{fa}$ and 
$M_{c_{k}}^{B}$. Thus, the ACA loss for category $c_{k}$ can be presented as
\begin{equation}
  \label{Laca_ck}
  \begin{split}
    \mathcal{L}_{aca}^{c_{k}} =&\max \left(\left(\varphi_{l} \times \mu_{c_{k}}^{ra} \right) -\mu_{c_{k}}^{fa},0 \right)+\\
    &\max \left(\left( \varphi_{g} \times \tau_{c_{k}}^{ra} \right) -\tau_{c_{k}}^{fa},0 \right), 
  \end{split} 
\end{equation}
where $\varphi_{l}$ and $\varphi_{g}$ are the thresholds for controlling the local (i.e., 
individual centroid features) and global (i.e., distribution of centroid features) similarity 
between the features of the synthesized image and the centroid features, respectively. At 
last, the ACA loss for all small sample categories can be shown as
\begin{equation}
  \label{L_aca}
  \mathcal{L}_{aca} =\frac{1}{N_{ssc}} \sum_{c_{k} \in \mathcal{C}_{ss}} \mathcal{L}_{aca}^{c_{k}},  
\end{equation}
where $N_{ssc}$ denotes the total number of small sample categories.

\subsection{Conditional Gradient Repair (CGR) Loss}
Although the colorization performance of small sample classes can be improved by memory-guided 
collaborative attention mechanisms, the problem of image gradient disappearance during 
translation remains under-addressed. To better deceive the patch discriminator, the generator 
usually smooths the intra-class edges (e.g., lane lines and window frames) or inter-class 
edges of the background class (e.g., tree and sky) as texture regions, which severely deviates 
from the scene layout of the original image. Therefore, a CGR loss is designed to selectively 
preserve the gradient structure of the original image, which encourages the gradient value of 
the translated image to be not smaller than the gradient at the corresponding location of the 
original image. Considering the existence of noisy regions with small gradient values in the 
NTIR image, the CGR loss focuses only on the structural preservation of regions with 
relatively large gradients in background categories. 

Due to the differences in the 
gradient distribution between images, we divide the gradients into two parts according to 
a sample-specific threshold (i.e., mean gradient value of the given image) rather than 
a fixed threshold. Specifically, the gradient maps of 
the NTIR image and its corresponding fake DC image are defined as 
$GM^{rb} \in \mathbb{R}^{H\times W}$ and $GM^{fa} \in \mathbb{R}^{H\times W}$, respectively. 
Given the binary mask $M_{bg}^{B}  \in \mathbb{R}^{H\times W}$ of the background region 
of the NTIR image, the gradient map of the background region can be denoted as
\begin{equation}
  \label{GM_bg}
  GM_{bg}^{rb} =M_{bg}^{B}\odot GM^{rb}. 
\end{equation}
Similarly, we obtain the gradient map of the background region corresponding to the 
translated image, denoted as $GM_{bg}^{fa}$. Then, we calculate the average gradient of 
the background region, denoted as $\rho$.
After that, we obtain the binary mask $M_{gh} \in \mathbb{R}^{H\times W}$ with a gradient 
greater than $\rho$. Finally, the CGR loss can be 
formulated as
\begin{equation}
  \label{L_cgr}
  \mathcal{L}_{cgr} =\frac{\sum relu\left( M_{gh}\odot \left( GM_{bg}^{rb} - GM_{bg}^{fa}\right)\right)}{\sum \left ( M_{gh}\odot GM_{bg}^{rb}\right)}, 
\end{equation}
where $relu\left( \cdot \right)$ denotes the rectified linear unit. With CGR loss, the 
structural consistency between the colorization result and the original image 
is further enhanced.

\subsection{Scale Robustness (SR) Loss}
\label{subsec_SR}
Inspired by self-supervised learning \cite{2020-ICML-Chen,2020-TNNLS-Zhao}, a SR loss 
is designed to 
improve the robustness of the model to variations in object scale, which encourages that the 
outputs corresponding to inputs of different scales can be resized to the same result. 
Concretely, taking domain A as an example, the inputs with $x_{a}$ scaled by factor 
$\alpha\left(< 1\right)$ and factor $\beta \left( > 1\right)$ are denoted as 
$x_{a}^{\alpha}$ and $x_{a}^{\beta}$, respectively. Then, 
$G_{AB}\left( x_{a} \right)$, $G_{AB}\left( x_{a}^{\alpha} \right)$, and 
$G_{AB}\left( x_{a}^{\beta} \right)$ are denoted as $O_{a}$, $O_{a}^{\alpha}$ and 
$O_{a}^{\beta}$, respectively. As the temperature differences between background categories 
in NTIR images are small, with reference to \cite{2022-TITS-Luo}, we use the smooth L1 
loss \cite{2017-CVPR-Zhu} $\mathcal{L}_{sl1}$ and SSIM \cite{2004-TIP-Wang} 
loss $\mathcal{L}_{ssim}$ to jointly capture the differences 
between images. Thus, given the inputs $x_{a}$ and $x_{a}^{\alpha}$, the corresponding SR loss 
can be expressed as
\begin{equation}
  \label{L_sr_A_a}
  \mathcal{L}_{sr}^{A_{\alpha}} = \lambda_{sl1} \mathcal{L}_{sl1}\left(O_{a}^{\alpha},O_{a}\downarrow_{\alpha} \right) +\mathcal{L}_{ssim}\left(O_{a}^{\alpha},O_{a}\downarrow_{\alpha}\right), 
\end{equation}
where $\lambda_{sl1}$ denotes the weight of smooth L1 loss, and $\downarrow_{\alpha}$ denotes 
down-sampling of $\alpha$ folds. Similarly, given the inputs $x_{a}$ and $x_{a}^{\beta}$, the 
corresponding SR loss, denoted as $\mathcal{L}_{sr}^{A_{\beta}}$, can be formulated as
\begin{equation}
  \label{L_sr_A_b}
  \mathcal{L}_{sr}^{A_{\beta}} = \lambda_{sl1}\mathcal{L}_{sl1} \left(O_{a}^{\beta}\downarrow_{\frac{1}{\beta}},O_{a}\right) +\mathcal{L}_{ssim} \left(O_{a}^{\beta}\downarrow_{\frac{1}{\beta}},O_{a}\right). 
\end{equation}
Further, the SR loss of domain A can be expressed as
\begin{equation}
  \label{L_sr_A}
  \mathcal{L}_{sr}^{A}=\mathbb{I} \left\{\psi < 0.5\right\}\mathcal{L}_{sr}^{A_{\alpha}}+\mathbb{I}\left\{ \psi \ge 0.5\right\}\mathcal{L}_{sr}^{A_{\beta}}, 
\end{equation}
where $\psi$ denotes a random variable within $\left [ 0,1 \right]$ varying with 
epoch. Similarly, we can obtain the SR loss $\mathcal{L}_{sr}^{B}$ for domain B. Finally, the 
total SR loss $\mathcal{L}_{sr}$ is the sum of $\mathcal{L}_{sr}^{A}$ and $\mathcal{L}_{sr}^{B}$.
With the introduction of SR loss, the problem of intra-class semantic inconsistency of large 
scale objects in translation results is further reduced.

\subsection{Objective Function}
In summary, the overall objective function of the proposed MornGAN can be expressed as:
\begin{equation}
  \label{L_all}
  \begin{split}
    \mathcal{L}_{all} =&\mathcal{L}_{adv}+\mathcal{L}_{cyc}+\lambda_{tv}\mathcal{L}_{tv}+\lambda_{sga}\mathcal{L}_{sga}\\
    &+\mathcal{L}_{seg}^{all}+\mathcal{L}_{aca}+\mathcal{L}_{cgr}+\mathcal{L}_{sr}, 
  \end{split} 
\end{equation}
where $\lambda_{tv}$ and $\lambda_{sga}$ denote the weights of the corresponding losses. 
Referring to \cite{2022-TITS-Luo}, $\lambda_{tv}$ and $\lambda_{sga}$ are set to 5 and 0.5, 
respectively. Referring to \cite{2017-CVPR-Zhu}, $\lambda_{sl1}$ in Eq. (\ref{L_sr_A_a}) is 
set to 10.

\section{Experiments}
In this section, we first introduce the datasets and evaluation metrics associated with the 
NTIR2DC task. Then, we describe the experimental settings and implementation details. 
Experimental results on the FLIR and KAIST datasets are then presented. Afterward, we 
perform an ablation analysis of the proposed modules, losses, and strategies. 
The discussion of the experiments is provided at the end.

\subsection{Datasets and Evaluation Metrics}
\subsubsection{Datasets}
The FLIR and KAIST datasets are two commonly used benchmarks for the NTIR2DC task. The FLIR 
Thermal Starter Dataset \cite{2019-FLIR-FA} provides TIR images with bounding box annotations 
for training and validation of the object detection model, while the reference RGB images are 
unannotated. Through the same data split as in \cite{2022-TITS-Luo}, we finally obtain 5447 DC 
images and 2899 NTIR images for training, while an additional 490 
NTIR images are used for testing.

The KAIST Multispectral Pedestrian Detection Benchmark \cite{2015-CVPR-Hwang} provides 
coarse-aligned RGB and TIR image pairs, which contain both daytime and nighttime conditions. 
Folldue \cite{2022-TITS-Luo}, the training set contains 1674 enhanced DC images and 1359 
NTIR images, and an additional 500 NTIR images are used as the test set to evaluate the 
semantic and edge consistency. For the pedestrian detection experiments, the sample size of 
the test set is 611.

In order to remove the black areas on both sides in some images, according to 
\cite{2022-TITS-Luo}, we first resize the training images to a resolution 
of $500\times 400$, and then the $360\times 288$ resolution images obtained by center 
cropping are used as training data.

\subsubsection{Evaluation Metrics}
To evaluate the performance for image content 
preservation at each level, we conduct experiments on three 
vision tasks: semantic segmentation, object detection, and edge preservation.

Intersection-over-Union (IoU) \cite{2015-IJCV-Everingham} is a widely used metric in 
semantic segmentation tasks. The mean value of IoU 
for all classes, denoted as mIoU, is adopted to evaluate the semantic consistency of NTIR 
image colorization methods.

Average precision (AP) \cite{2010-IJCV-Everingham} denotes the average detection 
precision of the object detection model under different recalls. The mean value of AP for 
all categories, defined as mAP, is selected as an overall evaluation metric.

APCE \cite{2022-TITS-Luo} is the average precision of Canny edges under multi-threshold 
conditions, and is employed to evaluate the edge preservation performance of the NTIR2DC model.

\subsection{Experimental Settings and Implementation Details}
We compare MornGAN with other NTIR2DC methods such as PearlGAN \cite{2022-TITS-Luo} 
and DlamGAN \cite{2021-ICIG-Luo}, as well as some low-light enhancement methods (e.g., 
ToDayGAN \cite{2019-ICRA-Anoosheh} and ForkGAN \cite{2020-ECCV-Zheng}) and prevalent I2I 
translation methods (e.g., CycleGAN \cite{2017-CVPR-Zhu}, UNIT \cite{2017-NIPS-Liu} and 
DRIT++ \cite{2020-IJCV-Lee}). We follow the instructions of these methods in order to establish a 
fair setting for comparison.

MornGAN is implemented using PyTorch. We train the models using the 
Adam \cite{2014-Arxiv-Kingma} optimizer 
with $\left(\beta_{1},\beta_{2}\right) =\left( 0.5,0.999 \right)$ on NVIDIA RTX 3090 GPUs. The
batch size is set to 1 for all experiments. The learning rate of the whole training process 
is maintained at 0.0002. The total number of training epochs for the FLIR and KAIST datasets 
are 80 and 160, respectively. In Eq. (\ref{th_LB}), the pseudo-label thresholds $\theta_{fg}$ 
and $\theta_{bg}$ are empirically set to 0.95 and 0.99, respectively, and $\mathcal{C}_{fg}$ 
includes buildings and all object categories, while the remaining categories all belong 
to $\mathcal{C}_{bg}$. In subsection \ref{subsec_MCA}, $\mathcal{C}_{ss}$ includes six 
categories: traffic light, traffic sign, person, truck, bus, and motorcycle. 
The number of clusters (i.e., $N_{u}$) for Kmeans clustering in ACA loss is set to four. In 
Eq. (\ref{Laca_ck}), the similarity thresholds $\varphi_{l}$ and $\varphi_{g}$ are both set 
to 0.9 to enhance the feature similarity within the classes. In subsection \ref{subsec_SR}, 
the scale factors $\alpha$ and $\beta$ are set to 0.5 and 1.5, respectively, to reduce the 
sensitivity of the model to object scale. For data augmentation, we flip the images 
horizontally with a probability of 0.5, and randomly crop them to $256\times 256$. 
The number of parameters of our model is about 46.7 MB, and the inference speed on an NVIDIA 
RTX 3090 GPU is about 0.01 seconds for an input image with a resolution of $360\times288$ pixels.

To achieve semantic consistency in translation, similar to DlamGAN \cite{2021-ICIG-Luo}, we 
divide the training process corresponding to segmentation loss into three 
phases\footnote{See \url{https://github.com/FuyaLuo/MornGAN/} for specific implementation 
details.}: 
learning $S_{A}$, learning $S_{B}$, and constraining semantic consistency 
after fixing $S_{A}$ and $S_{B}$. 

Due to the lack of pixel-level annotations in the FLIR and KAIST datasets, we evaluate the 
semantic segmentation performance of translated images using a scene parsing 
model \cite{2020-Arxiv-Tao} trained on Cityscape \cite{2016-CVPR-Cordts}, which considers both the 
feature plausibility and semantic consistency of colorization. Similarly, to measure the 
naturalness of object features, we utilize YOLOv4 \cite{2020-Arxiv-Bochkovskiy}, which is 
trained on the MS COCO \cite{2014-ECCV-Lin} dataset as the evaluation model for object detection.

\begin{figure}[!t]
\centering
\includegraphics[width=3.45in]{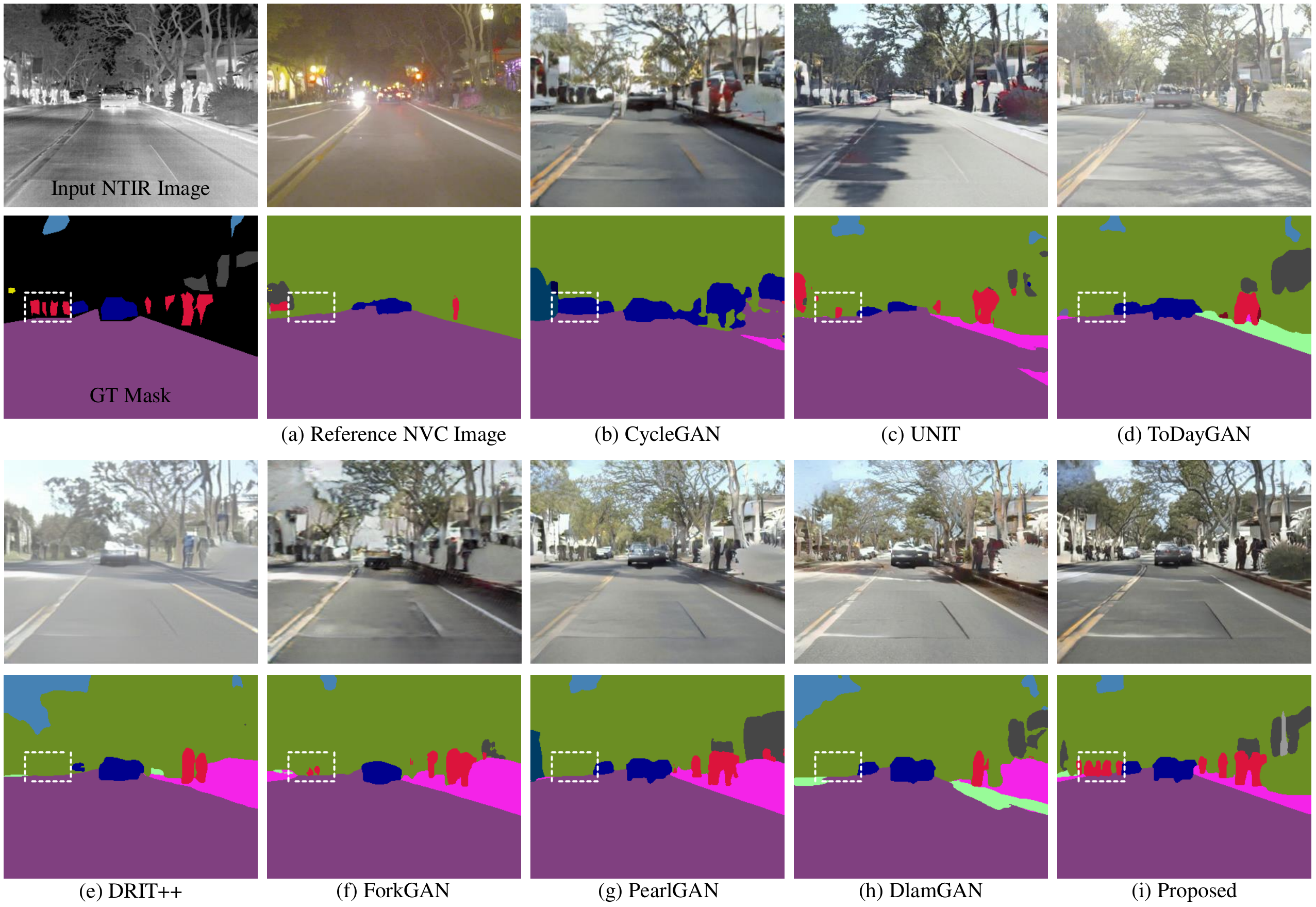}
\caption{The visual comparison of translation (the first row) and segmentation results (the second row) 
for different methods on the FLIR dataset. Please zoom in for more details on the 
content and quality. The areas in the white dotted boxes deserve attention.}
\label{fig_seg_flir}
\end{figure}

\begin{table*}[htbp]
  \centering
  \caption{Semantic segmentation performance (IoU) on the translated images by different 
  translation methods on the FLIR dataset.}
    \begin{tabular}{ccccccccccc}
      \toprule
          & Road  & Building & Sky   & Car   & Traffic Sign & Pedestrian & Motorcycle & Truck & Bus   & mIoU \\ \hline
    Reference NVC image & 95.2  & 53.7  & 1.0   & 56.6  & 5.2   & 40.3  & 0.0   & 2.5   & 70.1  & 36.1  \\
    CycleGAN \cite{2017-CVPR-Zhu} & 97.2  & 19.6  & 89.4  & 79.3  & 0.0   & 67.1  & 0.0   & 0.3   & 0.8   & 39.3  \\
    UNIT \cite{2017-NIPS-Liu}  & 96.3  & 48.3  & 92.5  & 63.7  & 0.0   & 59.5  & 12.4  & 0.6   & \textbf{49.4} & 47.0  \\
    ToDayGAN \cite{2019-ICRA-Anoosheh} & 97.0  & 42.3  & 83.2  & 76.5  & 0.0   & 56.3  & 2.5   & 0.0   & 6.5   & 40.5  \\
    DRIT++ \cite{2020-IJCV-Lee} & 98.2  & 16.1  & 75.3  & 79.4  & 0.0   & 38.3  & 0.0   & 0.0   & 3.2   & 34.5  \\
    ForkGAN \cite{2020-ECCV-Zheng} & 96.2  & 48.9  & 90.2  & 82.1  & 0.0   & 73.2  & 7.9   & 0.0   & 17.8  & 46.3  \\
    PearlGAN \cite{2022-TITS-Luo} & 98.6  & 71.0  & 95.1  & 89.1  & 0.0   & 84.3  & 12.2  & 1.5   & 0.0   & 50.2  \\
    DlamGAN \cite{2021-ICIG-Luo} & 97.4  & 67.2  & 94.0  & 89.1  & 0.1   & 74.0  & 36.2  & \textbf{2.1} & 0.3   & 51.2  \\
    Proposed & \textbf{98.7} & \textbf{84.0} & \textbf{96.8} & \textbf{95.6} & \textbf{0.7} & \textbf{94.5} & \textbf{40.4} & 1.2   & 22.8  & \textbf{59.4} \\
    \bottomrule
    \end{tabular}%
  \label{tab_flir_seg}%
\end{table*}%

\subsection{Experiments on the FLIR Dataset}
\subsubsection{Semantic Segmentation}
The translation results and corresponding semantic outputs of various methods are shown in 
Fig. \ref{fig_seg_flir}. Column (a) represents the reference nighttime visible color (NVC) 
image and its 
semantic segmentation. The segmentation model fails to discriminate the pedestrians on 
the side of the car due to the bright beam of car headlights and low surrounding illumination. 
As shown in the 
white dashed boxes in the second row, all the compared I2I translation methods fail to 
generate plausible pedestrians. In 
contrast, the proposed model can maintain the complete pedestrian region and pose to 
facilitate the segmentation model's discrimination, whether it covers a crowd (e.g., left side of 
the road) or an isolated pedestrian (e.g., right side of the road). Furthermore, the proposed 
approach outperforms other compared approaches for the structural preservation of the 
unlabeled sidewalk category (i.e., the pink region in the semantic mask). 

Table \ref{tab_flir_seg} reports the 
quantitative comparison of the semantic consistency of various translation models. 
The proposed MornGAN outperforms other methods in terms of semantic retention in four large 
sample categories (i.e., road, building, sky, and car) and three small sample categories (i.e., 
traffic sign, pedestrian, and motorcycle). As there are few available samples and diverse colors, 
all methods have poor translation performance for the traffic sign, truck, bus, and motorcycle 
categories. Benefiting from a memory-guided collaborative attention strategy, the proposed 
method slightly outperforms other methods in semantic preservation for 
traffic signs and motorcycles. However, due to the lack of spatial continuity constraints on 
large object representation, the proposed method produces limited improvement in translation 
performance for truck and bus. Overall, the proposed method outperforms the other methods by 
a significant margin (i.e., at least 8.2\%) in terms of mIoU for scene layout maintenance.

\subsubsection{Object Detection}
\begin{figure*}[!t]
\centering
\includegraphics[width=1\textwidth]{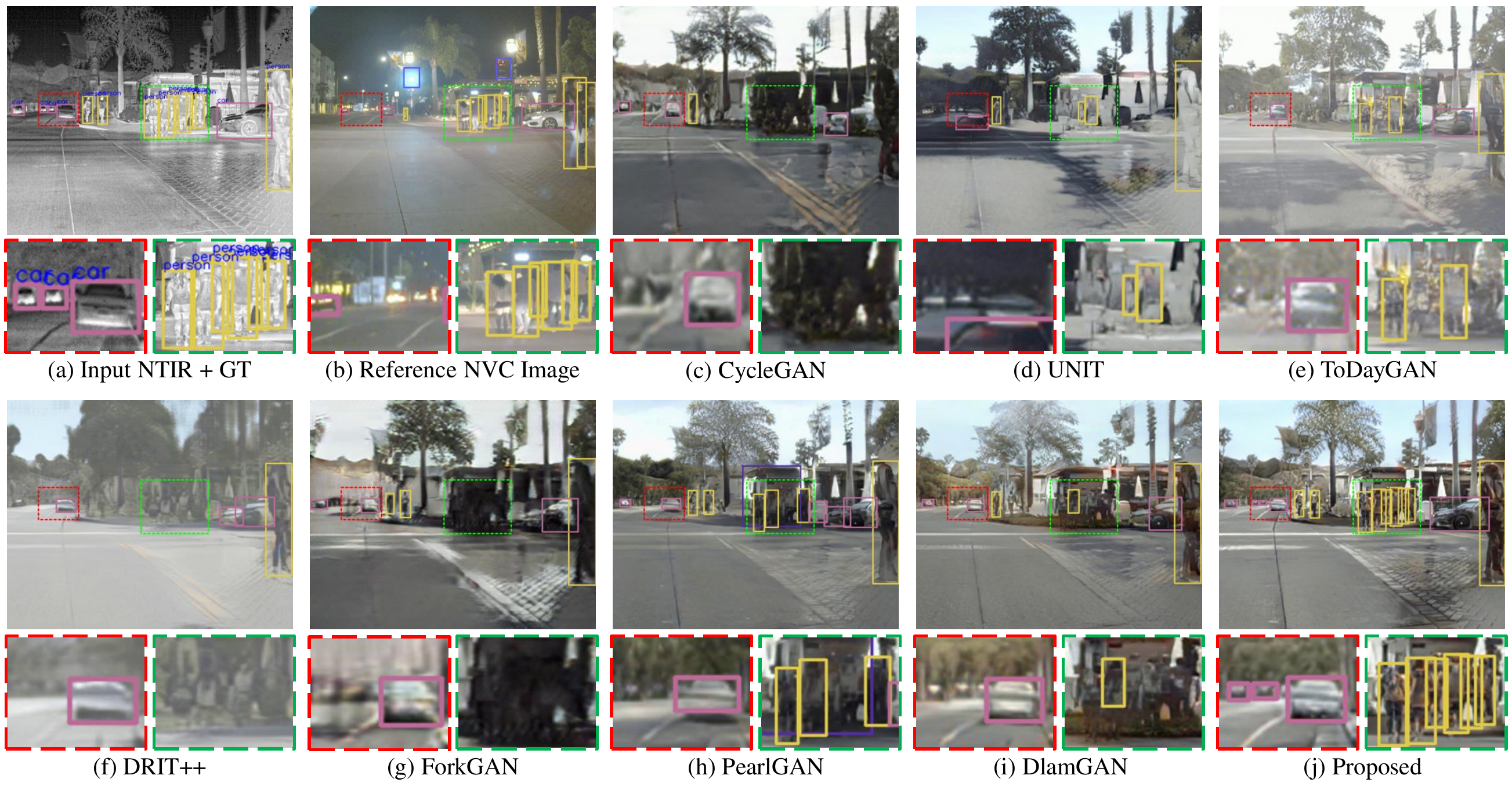}
\caption{Visual comparison of detection results on the FLIR dataset by YOLOv4 model \cite{2020-Arxiv-Bochkovskiy}. The parts covered by red and 
green dashed boxes show the enlarged patches in the corresponding images. Colors in the 
detection results that do not intersect with GT represent undefined categories of the FLIR 
dataset as identified by the detector.}
\label{fig_det_flir}
\end{figure*}

Better translation should facilitate better object detection. Fig. \ref{fig_det_flir} presents 
the qualitative comparison of the colorization and object detection by 
YOLOv4 \cite{2020-Arxiv-Bochkovskiy} on the translated images 
of various I2I translation methods. As shown in the red dashed boxes, almost all methods 
fail to reasonably 
translate the distant car except ours, which demonstrates the superiority of the proposed 
method for small object preservation. For the translation of the 
occluded objects, as shown in the green dashed boxes, all the I2I translation methods fail to 
make YOLOv4 recognize the complete six pedestrians except ours. 
Although YOLOv4 can identify pedestrians in well-lit areas of the NVC image, it 
cannot identify pedestrians in low-light areas (i.e., the area between the red dashed box 
and the green dashed box in the original image), which can be complemented by the proposed 
method. 

As the bounding-box annotation of the FLIR dataset covers only three categories 
(i.e., pedestrian, bicycle, and car), quantitative comparison of the various colorization 
results for object detection is shown in Table \ref{tab_flir_det}. The proposed method 
outperforms the other 
methods by a clear margin in object preservation for all categories. For example, 
MornGAN outperforms the second-ranked PearlGAN by a significant margin of 13.1\%, 
which indicates the superiority of our method in object retention.

\begin{table}[htbp]
  \centering
  \caption{Object detection performance (AP) on the translated images by different 
  translation methods on the FLIR dataset, computed at a single IoU of 0.50.}
    \begin{tabular}{ccccc} \toprule
          & Pedestrian & Bicycle & Car   & mAP \\ \hline
    Reference NVC image & 9.8   & 2.6   & 11.5  & 8.0  \\ 
    CycleGAN \cite{2017-CVPR-Zhu} & 17.8  & 1.9   & 37.2  & 19.0  \\
    UNIT \cite{2017-NIPS-Liu}  & 16.3  & 9.5   & 18.3  & 14.7  \\
    ToDayGAN \cite{2019-ICRA-Anoosheh} & 19.0  & 1.5   & 53.3  & 24.6  \\
    DRIT++ \cite{2020-IJCV-Lee} & 16.5  & 2.2   & 46.0  & 21.6  \\
    ForkGAN \cite{2020-ECCV-Zheng} & 25.9  & 2.3   & 32.5  & 20.2  \\
    PearlGAN \cite{2022-TITS-Luo} & 54.0  & 23.0  & 75.5  & 50.8  \\
    DlamGAN \cite{2021-ICIG-Luo} & 48.0  & 17.8  & 70.2  & 45.4  \\
    Proposed & \textbf{79.5} & \textbf{29.5} & \textbf{82.9} & \textbf{63.9} \\
    \bottomrule
    \end{tabular}%
  \label{tab_flir_det}%
\end{table}%

\subsubsection{Edge Preservation}
\begin{figure*}[!t]
\centering
\includegraphics[width=1\textwidth]{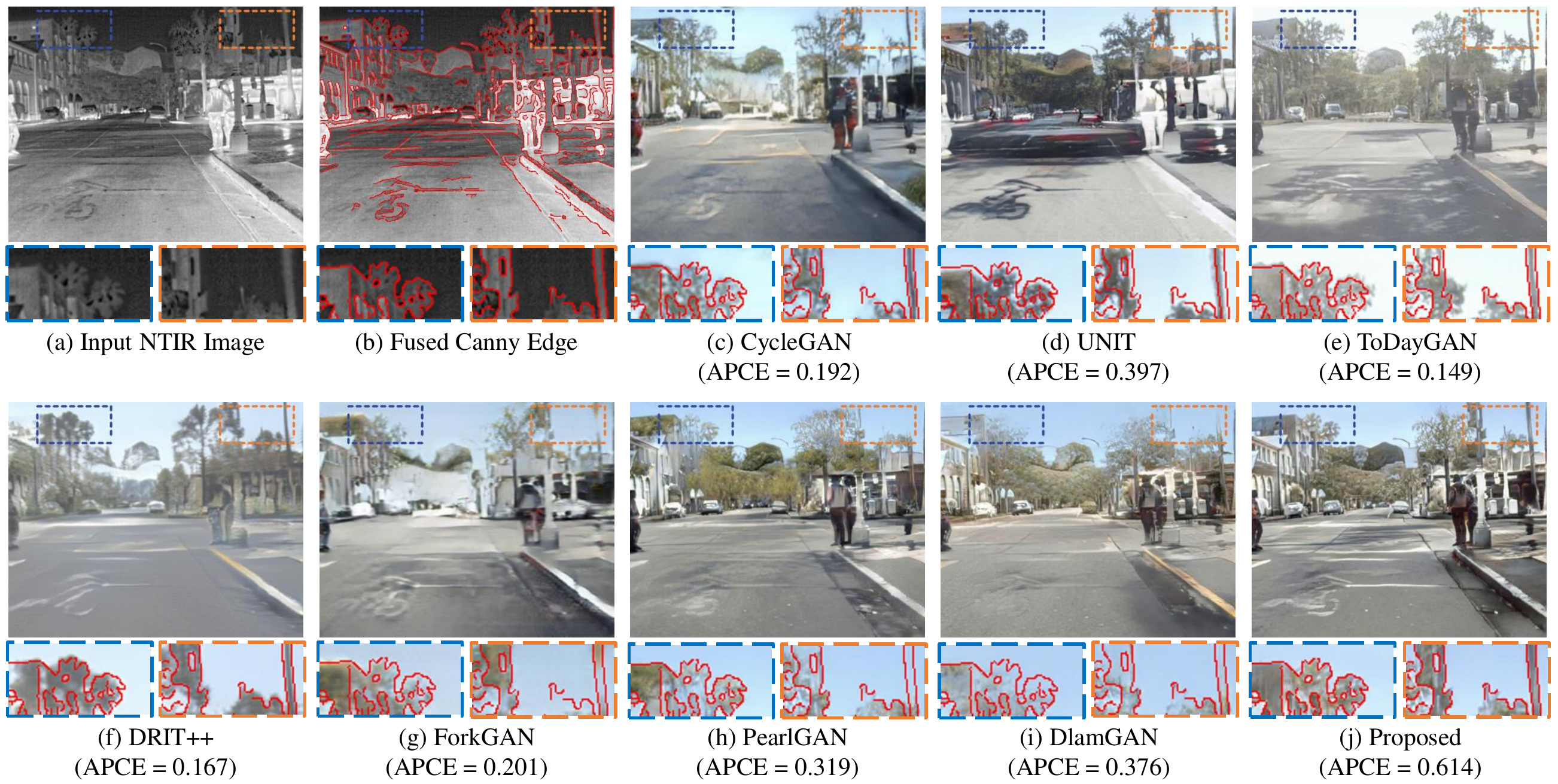}
\caption{Visual comparison of geometric consistency on the FLIR dataset. The second row 
shows the enlarged results of the corresponding regions after fusion with the edges. 
The edges in red are extracted by the Canny detector from the input NTIR image.}
\label{fig_edge_flir}
\end{figure*}

Fig. \ref{fig_edge_flir} visually compares the edge consistency of various translation 
methods. As shown in 
the blue dashed boxes, the edges of the buildings in the results of CycleGAN, ToDayGAN and 
DRIT++ are outwardly expanded, while the edges of the trees in the other four compared 
methods are inwardly shrunken. On the contrary, our model provides a complete match to the edges 
of the original image. In addition, as shown in the orange dashed boxes, ForkGAN, PearlGAN, and 
DlamGAN fail to maintain the continuous structure of the pole, and the edges of the pole are 
inwardly contracted in the other four methods. Compared with the other methods, 
MornGAN can more faithfully adhere to the edge structure of the original image. 

\begin{figure}[!t]
\centering
\includegraphics[width=3.45in]{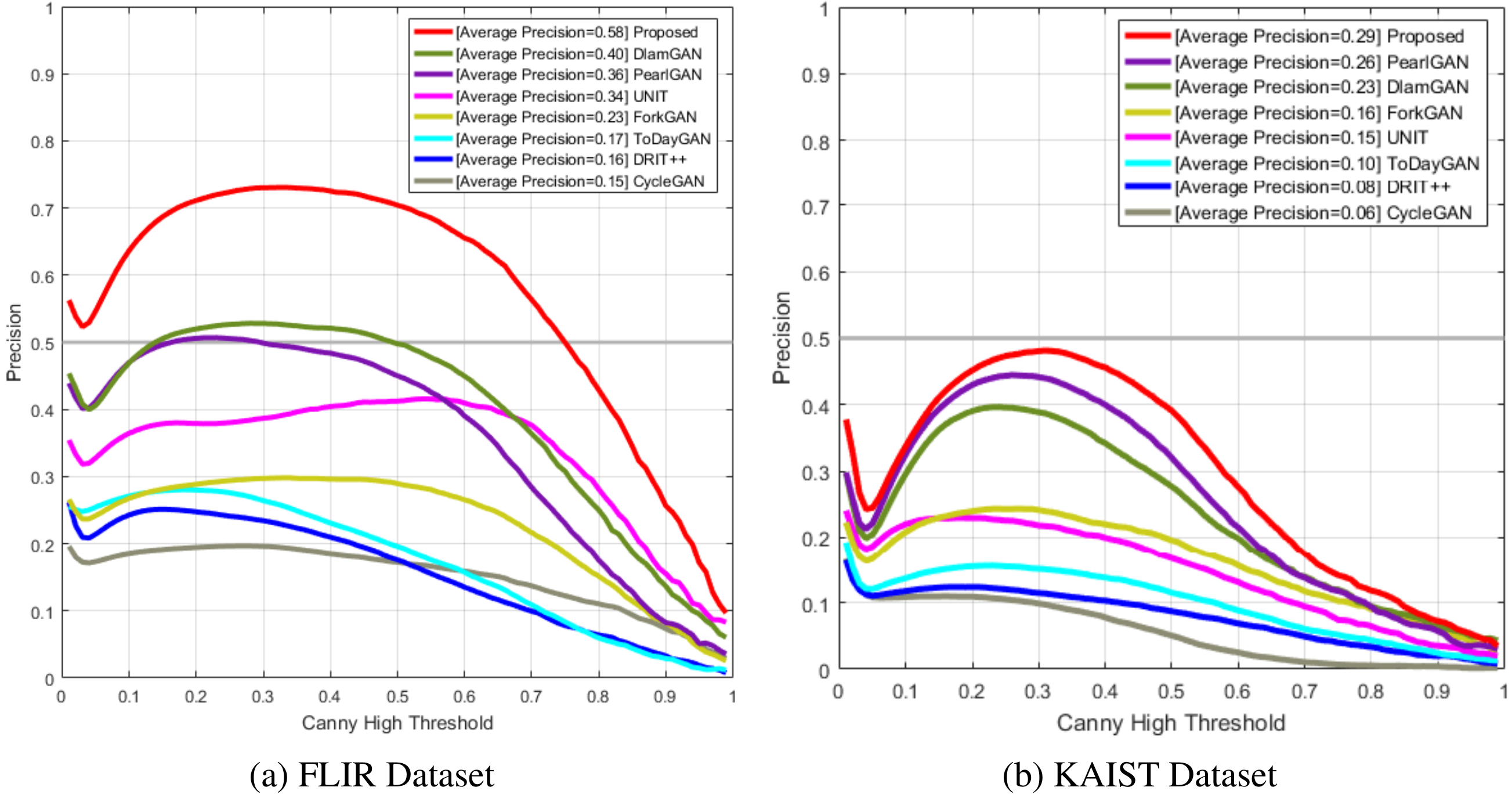}
\caption{APCE results of different translation methods on the FLIR and KAIST datasets.}
\label{fig_apce}
\end{figure}

As the Canny edges in the Fig. \ref{fig_edge_flir} are only the results of a fixed threshold, 
we exploit the 
APCE metric, which covers multiple thresholds to comprehensively evaluate the edge consistency 
performance, as shown in Fig. \ref{fig_apce}(a). We can find that the proposed method significantly 
outperforms other methods in edge consistency at all thresholds and is far 
superior to the second ranked DlamGAN by 18\%.

\subsection{Experiments on the KAIST Dataset}
Different from FLIR \cite{2019-FLIR-FA}, KAIST \cite{2015-CVPR-Hwang} is a more challenging 
dataset with low-contrast and blurred NTIR images.

\begin{table*}[htbp]
  \centering
  \caption{Semantic segmentation performance (IoU) on the translated images by different 
  translation methods on the KAIST dataset.}
    \begin{tabular}{cccccccccc} \toprule
          & Road  & Building & Sky   & Car   & Traffic Sign & Pedestrian & Motorcycle & Bus   & mIoU \\ \hline
    Reference NVC image & 82.6  & 74.6  & 0.0   & 56.1  & 0.0   & 19.0  & 0.0   & 0.0   & 29.0  \\
    CycleGAN \cite{2017-CVPR-Zhu} & 83.5  & 25.0  & 64.1  & 39.4  & 0.0   & 8.9   & 0.1   & 0.0   & 27.6  \\
    UNIT \cite{2017-NIPS-Liu}  & 92.3  & 59.3  & 81.0  & 60.3  & 0.0   & 23.8  & 0.0   & 0.0   & 39.6  \\
    ToDayGAN \cite{2019-ICRA-Anoosheh} & 92.5  & 57.6  & 82.8  & 54.1  & 3.6   & 23.3  & 0.1   & 0.0   & 39.3  \\
    DRIT++ \cite{2020-IJCV-Lee} & 89.1  & 58.9  & 63.1  & 47.6  & 0.0   & 5.6   & 0.0   & 0.0   & 33.0  \\
    ForkGAN \cite{2020-ECCV-Zheng} & 89.6  & 30.3  & 44.6  & 48.5  & 0.5   & 27.9  & 0.0   & 0.0   & 30.2  \\
    PearlGAN \cite{2022-TITS-Luo} & 93.4  & 43.1  & 83.8  & 70.3  & 0.9   & 57.6  & 0.0   & 6.1   & 44.4  \\
    DlamGAN \cite{2021-ICIG-Luo} & 92.6  & 49.0  & 71.9  & 66.4  & 2.1   & 49.5  & 2.0   & 8.2   & 42.7  \\
    Proposed & \textbf{93.7} & \textbf{72.2} & \textbf{87.6} & \textbf{72.5} & \textbf{7.2} & \textbf{61.1} & \textbf{4.2} & \textbf{13.1} & \textbf{51.5} \\
    \bottomrule  
  \end{tabular}%
  \label{tab_kaist_seg}%
\end{table*}%

\subsubsection{Semantic Segmentation}
\begin{figure}[!t]
\centering
\includegraphics[width=3.45in]{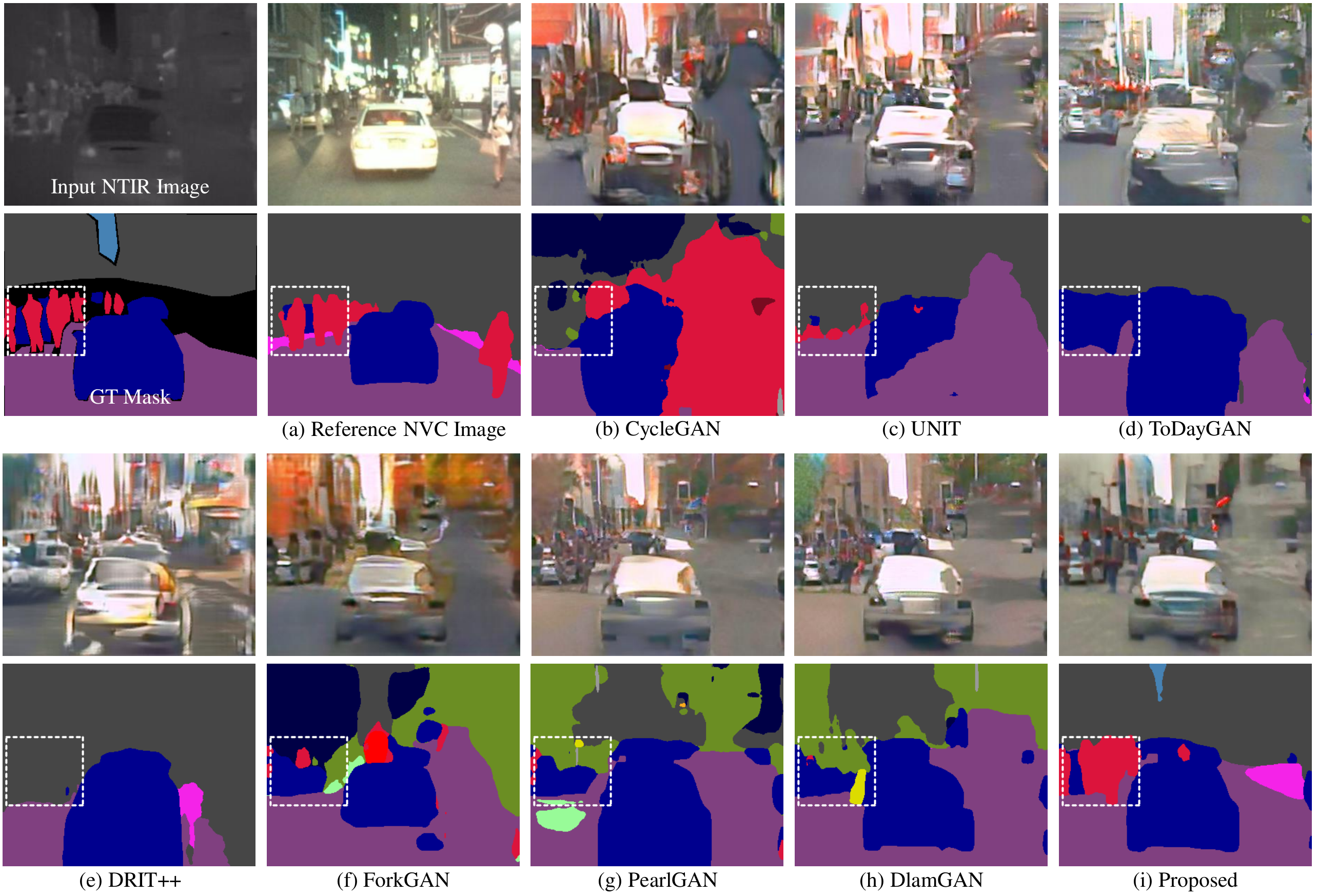}
\caption{The visual comparison of translation (the first row) and segmentation performance 
(the second row) of different methods on the KAIST dataset. The areas in white dotted boxes 
deserve attention.}
\label{fig_seg_kaist}
\end{figure}

Fig. \ref{fig_seg_kaist} visually compares the colorization results of various translation 
methods on the KAIST 
dataset and their segmentation outputs. The relatively reasonable 
segmentation results in column (a) demonstrate the applicability of the selected segmentation 
model proposed in \cite{2020-Arxiv-Tao}. However, as shown in the white dashed boxes, all the 
compared I2I translation methods 
are unable to generate realistic pedestrians for the segmentation model to discriminate. 
Instead, the proposed method not only maintains most of the semantics of the dashed box 
region but also provides partial clues for distant pedestrian detection.

Further, a quantitative comparison of the semantic preservation performance 
of various I2I translation methods is shown in Table \ref{tab_kaist_seg}. Despite the poor image 
quality, which 
makes scene understanding extremely difficult, MornGAN achieves the best 
performance among all methods in terms of semantic preservation for each category. Similar 
to the results on the FLIR dataset, all I2I translation methods have poor semantic retention 
in small sample categories. With the help of the proposed learning approach, features 
of small sample categories can be better retained compared with other methods. Overall, the 
semantic consistency of MornGAN on the NTIR2DC task is far superior to that of other 
I2I translation methods, that is, at least 7.1\% ahead.

\subsubsection{Pedestrian Detection}
\begin{figure*}[!t]
\centering
\includegraphics[width=1\textwidth]{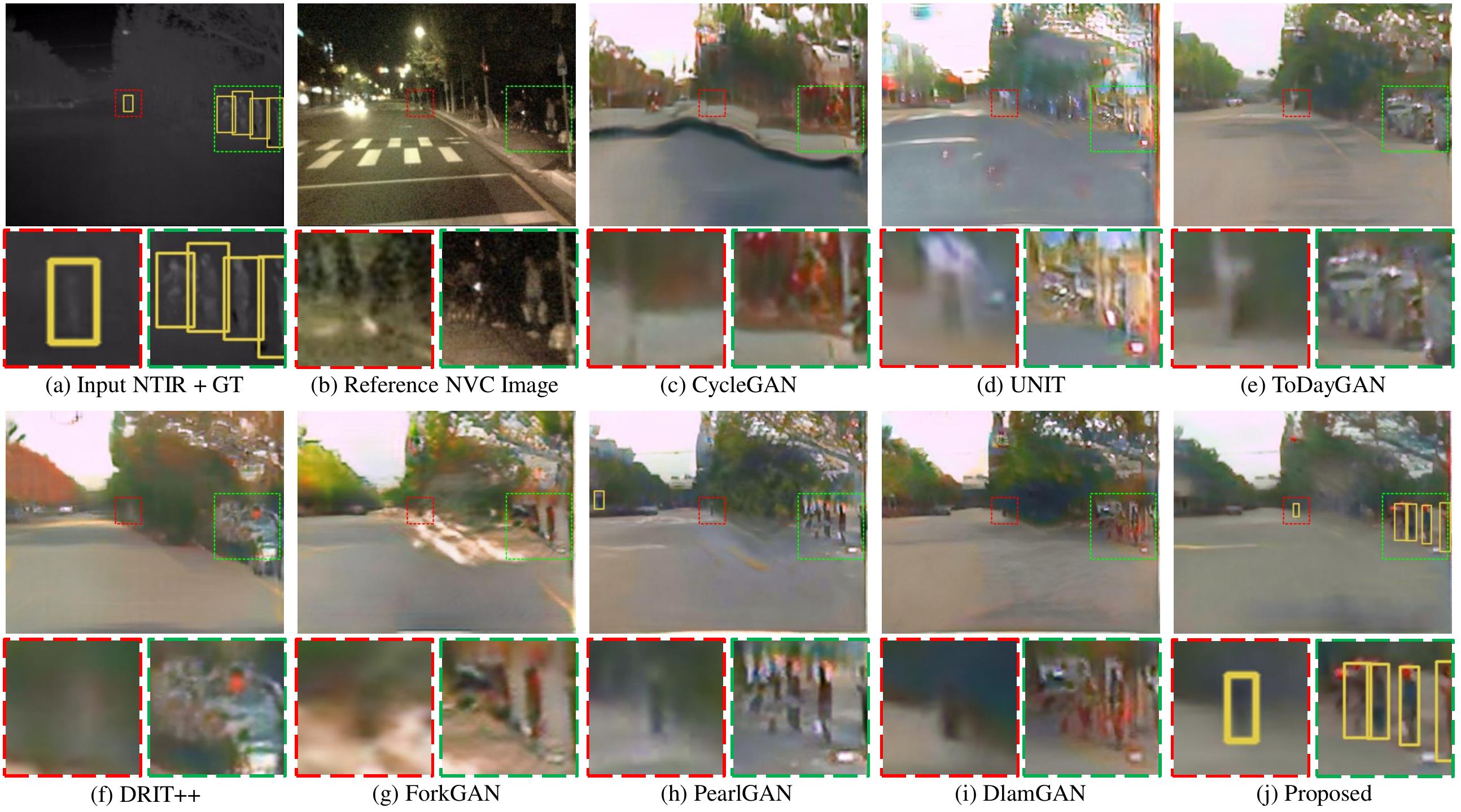}
\caption{Visual comparison of pedestrian detection by the YOLOv4 model \cite{2020-Arxiv-Bochkovskiy} 
on the KAIST dataset.}
\label{fig_det_kaist}
\end{figure*}

For pedestrian preservation, a qualitative comparison of various translation methods on the 
KAIST dataset is shown in Fig. \ref{fig_det_kaist}. As shown in 
the two dashed boxes, all the methods cannot maintain realistic and complete pedestrian 
features to make the detection model convincing. However, MornGAN can generate 
plausible features for near pedestrians while maintaining a relatively reasonable local 
feature distribution for distant pedestrians. 

Further, the quantitative comparison for pedestrian preservation is shown in 
Table \ref{tab_kaist_det}. Due to the large number of pedestrians and good 
lighting conditions in city scenes, the performance of pedestrian detection in the reference 
NVC images is slightly better than that of the proposed method. Nevertheless, MornGAN 
outperforms other methods both in terms of detection precision and recall, and has substantial 
advantages in terms of overall mAP metrics.

\begin{table}[htbp]
  \centering
  \caption{Pedestrian detection performance on the translated images by different 
  translation methods on the KAIST dataset, computed at a single IoU of 0.50.}
    \begin{tabular}{cccc} \toprule
          & Precision & Recall & mAP \\ \hline
    Reference NVC image & 36.8  & 50.1  & 44.2  \\
    CycleGAN \cite{2017-CVPR-Zhu} & 4.7   & 2.8   & 1.1  \\
    UNIT \cite{2017-NIPS-Liu}  & 26.7  & 14.5  & 11.0  \\
    ToDayGAN \cite{2019-ICRA-Anoosheh} & 11.4  & 14.9  & 5.0  \\
    DRIT++ \cite{2020-IJCV-Lee} & 7.9   & 4.1   & 1.2  \\
    ForkGAN \cite{2020-ECCV-Zheng} & 33.9  & 4.6   & 4.9  \\
    PearlGAN \cite{2022-TITS-Luo} & 21.0  & 39.8  & 25.8  \\
    DlamGAN \cite{2021-ICIG-Luo} & 26.1  & 32.0  & 23.0  \\
    Proposed & \textbf{34.3} & \textbf{52.7} & \textbf{42.6} \\
    \bottomrule  
  \end{tabular}%
  \label{tab_kaist_det}%
\end{table}%

\subsubsection{Edge Preservation}
\begin{figure*}[!t]
\centering
\includegraphics[width=1\textwidth]{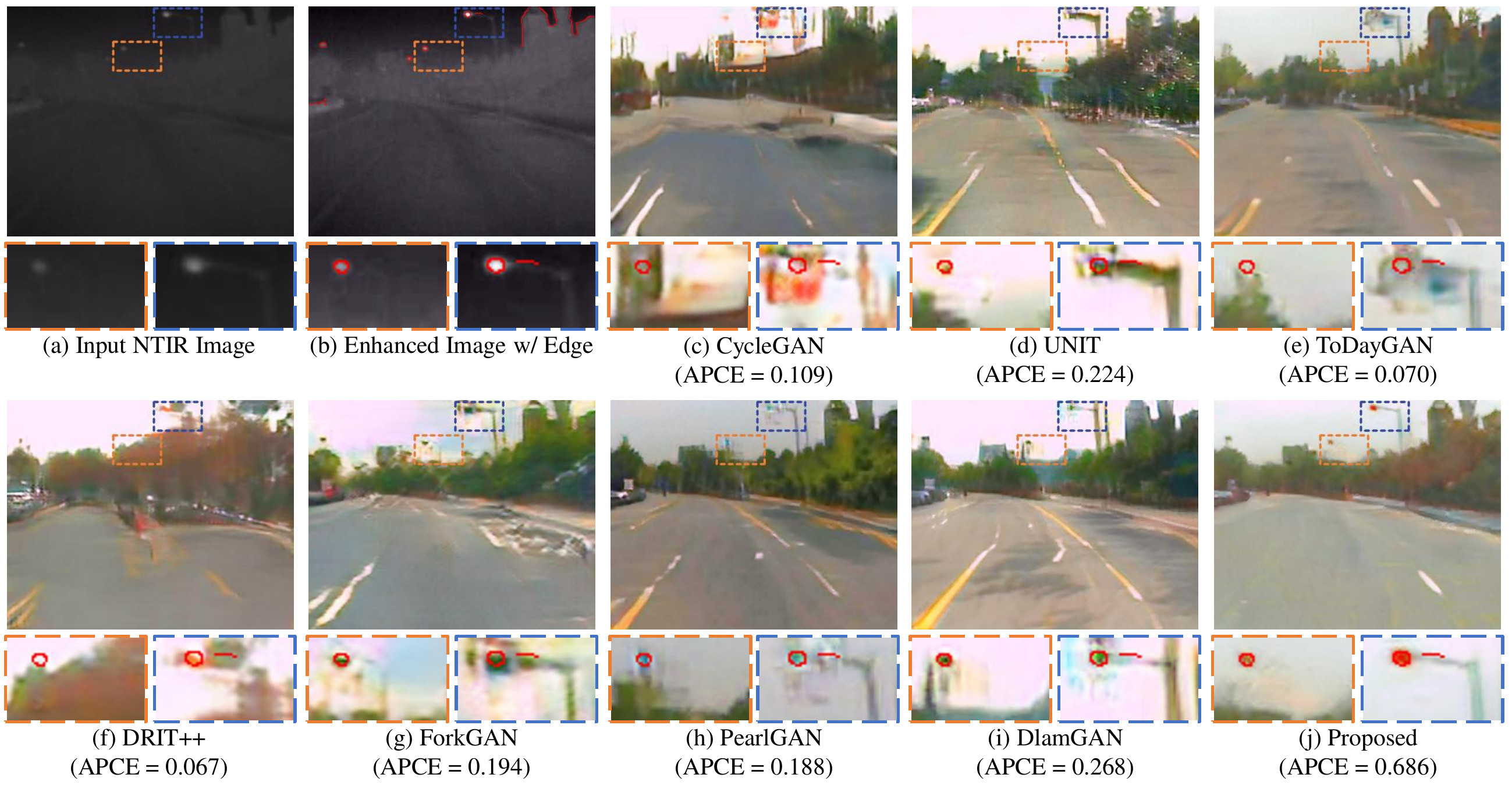}
\caption{Visual comparison of geometric consistency on the KAIST dataset. Column (b) 
shows the enhanced image of the input image overlapped with its 
Canny edges in red for better viewing.}
\label{fig_edge_kaist}
\end{figure*}

Fig. \ref{fig_edge_kaist} qualitatively compares the edge consistency of various 
translation methods on the KAIST dataset. Column (b) shows the enhanced NTIR 
image overlapped with Canny edges. As shown in the orange dashed boxes, the streetlights 
are vanishing in the results of 
ToDayGAN and DRIT++, while the edges of the streetlights are indented in the results of UNIT, 
ForkGAN, and PearlGAN. The neighboring structures of streetlights in CycleGAN and DlamGAN 
deviate severely from the input NTIR image. Similarly, as shown in the blue dashed boxes, the 
structures of the poles in the results of PearlGAN and DlamGAN are disconnected, while the 
streetlights and their neighboring edges of other methods differ significantly from the 
original image. Instead, our results match well with the original image on the 
edges. 

Furthermore, the edge consistency comparison under the multi-threshold condition is shown in 
Fig. \ref{fig_apce} (b). Considering all the thresholds, the proposed method still exhibits 
high performance in the edge consistency of the translation.

\begin{figure*}[!t]
\centering
\includegraphics[width=1\textwidth]{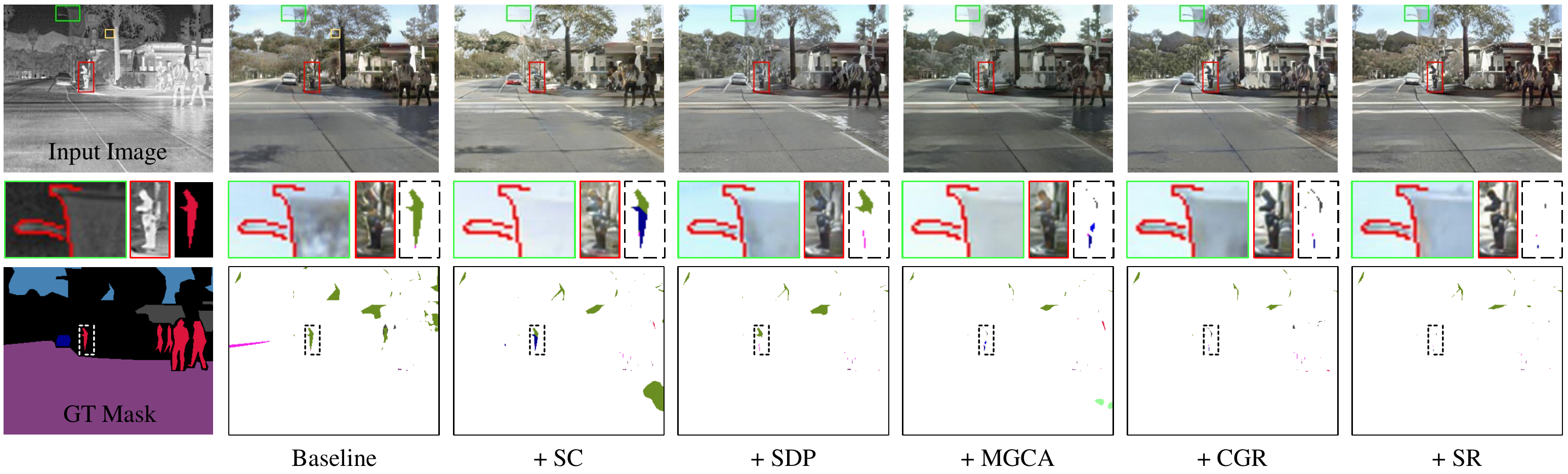}
\caption{Visual results of ablation study on the FLIR dataset. The first row shows the input NTIR 
image and the translated images by different models. In the second row, the parts covered by green 
boxes show the enlarged results of the corresponding regions after fusion with the 
edges of the input image. The parts covered by red boxes show the enlarged 
cropped regions in the corresponding image. The black dotted box is the result of zooming 
in on the corresponding area in the third row. The third row shows the error maps of the 
semantic segmentation results, where the white areas indicate the correct regions or 
unlabeled regions. The meanings of SC, SDP, MGCA, CGR, and SR can be found in Table \ref{tab_aa}.}
\label{fig_ablation}
\end{figure*}

\subsection{Ablation Study}
Ablation analysis is performed on the FLIR dataset to discuss the validity of each 
component of MornGAN. The results of the ablation analysis are shown in Table \ref{tab_aa}, and an 
example of a qualitative comparison is shown in Fig. \ref{fig_ablation}. We can 
find from Fig. \ref{fig_ablation} that the baseline 
model has poor content preservation performance for the input and tends to generate more 
erroneous regions of the tree (e.g., the yellow box region in the second column). With 
the introduction of semantic consistency loss, the 
semantic retention performance of the small-sample object class is degraded by the 
presence of excessive noise in the pseudo-labels, despite the improvement of the overall 
semantic consistency. Moreover, the wrong semantic labels lead to degradation of edge 
consistency, as shown in the green box in the third column in Fig. \ref{fig_ablation}. 

In the next experiment, we perform SDP on the pseudo-labels, and the semantic 
preservation as well as edge consistency of the results are improved, as shown in the 
fourth column of Fig. \ref{fig_ablation}. Then, due to the benefits of the MGCA strategy, 
the semantic 
preservation for the small sample category can be significantly improved, that is, 
a 3.4\% gain, as shown in Table \ref{tab_aa}. 

With the introduction of CGR loss, the edge structure of the street light is clearer 
compared to the previous results, as shown in the green box in the sixth column of 
Fig. \ref{fig_ablation}. The results in Table \ref{tab_aa} further demonstrate the 
effectiveness of CGR loss for improving edge consistency. 

Ultimately, as shown in the green box in the last column of Fig. \ref{fig_ablation}, 
the SR loss can facilitate the structure of the 
pole matching that of the original image. Moreover, as shown in the red box and the black dashed box, 
the proposed method can generate more plausible pedestrians and reduce the content distortion. 
Compared with the baseline model, the proposed MornGAN can substantially improve 
the performance for image content and edge preservation without increasing model parameters.

\begin{table}[htbp]
  \centering
  \caption{Quantitative ablation study on the FLIR dataset. ``SC" means the 
  semantic consistency loss. ``SDP" means the semantic denoising process. ``MGCA" means the 
  memory-guided collaborative attention. ``CGR" means the conditional gradient 
  repair loss. ``SR" means the scale robustness loss. ``mIoU.A" and ``mIoU.S" denote the mIoU 
  results for the set of all categories and small sample categories, respectively.}
  \renewcommand\tabcolsep{2.5pt}
    \begin{tabular}{ccccccccc} \toprule
    Baseline & SC    & SDP   & MGCA   & CGR   & SR    & mIoU.A(\%) & mIoU.S(\%) & APCE \\ \hline
    \checkmark     &       &       &       &       &       & 50.5  & 23.8  & 0.38  \\
    \checkmark     & \checkmark     &       &       &       &       & 51.8  & 20.5  & 0.37  \\
    \checkmark     & \checkmark     & \checkmark     &       &       &       & 54.0  & 23.6  & 0.39  \\
    \checkmark     & \checkmark     & \checkmark     & \checkmark     &       &       & 56.8  & 27.0  & 0.43  \\
    \checkmark     & \checkmark     & \checkmark     & \checkmark     & \checkmark     &       & 58.5  & 30.8  & 0.58  \\
    \checkmark     & \checkmark     & \checkmark     & \checkmark     & \checkmark     & \checkmark     & \textbf{59.4}  & \textbf{31.9}  & \textbf{0.58}  \\
    \bottomrule  
  \end{tabular}%
  \label{tab_aa}%
\end{table}%

\subsection{Discussion}
In this section, we analyze $\left( 1\right)$ the generalization capability of the proposed 
method, $\left( 2\right)$ the impact of the cluster number in the ACA loss on the results, and 
$\left( 3\right)$ the limitations of MornGAN.
\subsubsection{Generalization Experiments}
In order to explore the generalization ability of various I2I translation methods to 
out-of-domain distributions, we apply each model trained on the KAIST dataset to the FLIR 
dataset abbreviated as K$\rightarrow$F and vice versa as F$\rightarrow$K. The results are 
shown in Table \ref{tab_genexp}. For the performance of semantic 
preservation, we can find that the proposed method obtains the best mIoU among all 
methods for both K$\rightarrow$F and F$\rightarrow$K experiments. Similarly, for the 
comparison of edge consistency 
during translation, MornGAN outperforms other compared methods by a 
significant margin for APCE in both experimental paradigms. In summary, the results 
demonstrate the strong generalization capability of the proposed method for domain shift.

\begin{table}[htbp]
  \centering
  \caption{Results of generalization experiments. ``AVE.'' means average value.}
  \renewcommand\tabcolsep{2.5pt}
    \begin{tabular}{|c|c|c|c|c|c|c|}
      \hline
      & \multicolumn{3}{c|}{mIoU(\%)} & \multicolumn{3}{c|}{APCE} \\
      \cline{2-7}
      & K$\rightarrow$F   & F$\rightarrow$K   & AVE. & K$\rightarrow$F   & F$\rightarrow$K   & AVE. \\ \hline
      \hline
    CycleGAN \cite{2017-CVPR-Zhu} & 15.8  & 23.4  & 19.6  & 0.07  & 0.09  & 0.08  \\
    UNIT \cite{2017-NIPS-Liu}  & 23.9  & 19.3  & 21.6  & 0.18  & 0.22  & 0.20  \\
    ToDayGAN \cite{2019-ICRA-Anoosheh} & 27.9  & 18.5  & 23.2  & 0.09  & 0.10  & 0.10  \\
    DRIT++ \cite{2020-IJCV-Lee} & 12.5  & 23.0  & 17.8  & 0.06  & 0.14  & 0.10  \\
    ForkGAN \cite{2020-ECCV-Zheng} & 39.0  & 15.1  & 27.1  & 0.23  & 0.10  & 0.17  \\
    PearlGAN \cite{2022-TITS-Luo} & 37.8  & 28.5  & 33.2  & 0.21  & 0.24  & 0.23  \\
    DlamGAN \cite{2021-ICIG-Luo} & 32.4  & 20.2  & 26.3  & 0.18  & 0.30  & 0.24  \\
    Proposed & \textbf{39.9} & \textbf{41.6} & \textbf{40.8} & \textbf{0.29} & \textbf{0.45} & \textbf{0.37} \\
    \hline
  \end{tabular}%
  \label{tab_genexp}%
\end{table}%

\subsubsection{The Effect of Cluster Number on The Results}
The effect of the cluster number in ACA loss on the colorization performance is shown in 
Table \ref{tab_cluster}. When the number of clusters is small, the features of small 
sample objects are difficult to capture comprehensively, so the segmentation 
model fail to understand the presence of objects. However, when the number of clusters 
is too large, many irrelevant features in the objects may disturb the recognition of 
the segmentation model. Compared with semantic consistency, edge consistency during translation 
is less sensitive to the cluster number. Therefore, we set the cluster number to four 
in all experiments.

\begin{table}[htbp]
  \centering
  \caption{The impact of the cluster number in the ACA loss. ``mIoU.A" and ``mIoU.S" denote the mIoU 
  results for the set of all categories and small sample categories, respectively.}
    \begin{tabular}{|c|c|c|c|c|c|}
      \hline
    Cluster Number & 2     & 3     & 4     & 5     & 6 \\ \hline
    \hline
    mIoU.A (\%) & 57.1  & 58.6  & \textbf{59.4} & 55.5  & 56.7  \\
    mIoU.S (\%) & 28.0  & 30.2  & \textbf{31.9} & 26.1  & 26.2  \\
    APCE  & 0.56  & 0.58  & 0.58  & 0.57  & \textbf{0.59} \\
    \hline
    \end{tabular}%
  \label{tab_cluster}%
\end{table}%

\subsubsection{Failure Cases}
\begin{figure}[!t]
\centering
\includegraphics[width=2.8in]{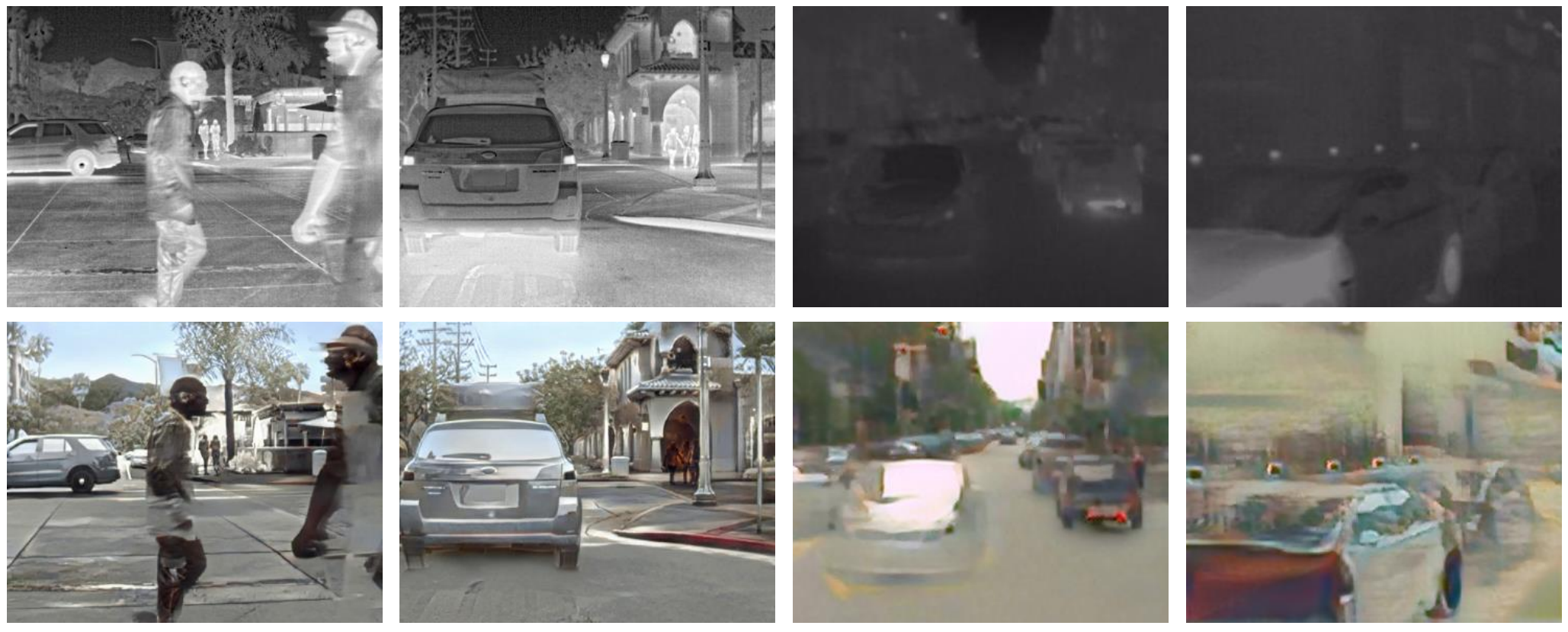}
\caption{Visualization of failure cases. The first and second rows show the NTIR images and 
their translation results, respectively.}
\label{fig_failure}
\end{figure}

Fig. \ref{fig_failure} shows four failure cases of MornGAN, where the input images in the first two columns 
are from the FLIR dataset, and the remaining images are from the KAIST dataset. As 
shown in the first and 
third columns, local areas of large scale pedestrians and cars are incorrectly translated as 
roads due to their similarity in temperature. Moreover, the model fails in generating plausible 
objects when the objects are close to the camera, as shown in the second and fourth columns. 
Therefore, more attempts should be made in the future to design reasonable modules to 
capture complete object representations, such as combining the integrity and continuity 
of Gestalt laws.

\section{Conclusion}
In this work, we developed a new learning framework called MornGAN to achieve colorization 
of NTIR images. Benefiting from the proposed memory-guided sample selection strategy and 
adaptive collaborative attention loss, the framework enabled the great improvement of the 
translation performance of small sample classes. The online semantic distillation module 
was designed to mine and refine 
the pseudo-labels of NTIR images. In addition, we devised a conditional gradient repair loss 
for avoiding image gradient disappearance during translation. Scale robustness loss was 
introduced to improve the robustness of the model to scale variation. Experiments on the 
NTIR2DC task demonstrated the superiority of the proposed approach in terms of semantic 
preservation and edge consistency, which remarkably improved the object detection performance 
on the translated images. Although the proposed semantic denoising process was only applied to 
a few categories with the low-rank property, its unsupervised and threshold-free advantages make it 
easy to extend to other tasks (e.g., saliency detection and image matting). In the 
future, it is a promising 
direction to further improve the reliability of scene parsing of NTIR images under 
weakly supervised conditions to ensure the semantic consistency of NTIR2DC tasks.

% use section* for acknowledgment
\ifCLASSOPTIONcompsoc
  % The Computer Society usually uses the plural form
  \section*{Acknowledgments}
\else
  % regular IEEE prefers the singular form
  \section*{Acknowledgment}
\fi

This work was supported by Key Area R\&D Program of Guangdong Province 
(2018B030338001) and National Natural Science Foundation of China (61806041, 62076055).

\ifCLASSOPTIONcaptionsoff
  \newpage
\fi

\bibliographystyle{IEEEtran}

\bibliography{refabrv}

% Generated by IEEEtran.bst, version: 1.13 (2008/09/30)
\begin{thebibliography}{10}
\providecommand{\url}[1]{#1}
\csname url@samestyle\endcsname
\providecommand{\newblock}{\relax}
\providecommand{\bibinfo}[2]{#2}
\providecommand{\BIBentrySTDinterwordspacing}{\spaceskip=0pt\relax}
\providecommand{\BIBentryALTinterwordstretchfactor}{4}
\providecommand{\BIBentryALTinterwordspacing}{\spaceskip=\fontdimen2\font plus
\BIBentryALTinterwordstretchfactor\fontdimen3\font minus
  \fontdimen4\font\relax}
\providecommand{\BIBforeignlanguage}[2]{{%
\expandafter\ifx\csname l@#1\endcsname\relax
\typeout{** WARNING: IEEEtran.bst: No hyphenation pattern has been}%
\typeout{** loaded for the language `#1'. Using the pattern for}%
\typeout{** the default language instead.}%
\else
\language=\csname l@#1\endcsname
\fi
#2}}
\providecommand{\BIBdecl}{\relax}
\BIBdecl

\bibitem{2009-EOJ-W}
G.~W~Stuart and P.~K~Hughes, ``Towards an understanding of the effect of night
  vision display imagery on scene recognition,'' \emph{The Ergonomics Open
  Journal}, vol.~2, no.~1, 2009.

\bibitem{2014-NIPS-Goodfellow}
I.~J. Goodfellow, J.~Pouget-Abadie, M.~Mirza, B.~Xu, D.~Warde-Farley, S.~Ozair,
  A.~C. Courville, and Y.~Bengio, ``Generative adversarial nets,'' in
  \emph{Proc. NeurIPS}, 2014.

\bibitem{2017-CVPR-Zhu}
J.-Y. Zhu, T.~Park, P.~Isola, and A.~A. Efros, ``Unpaired image-to-image
  translation using cycle-consistent adversarial networks,'' in \emph{Proc.
  ICCV}, 2017, pp. 2223--2232.

\bibitem{2017-NIPS-Liu}
M.-Y. Liu, T.~Breuel, and J.~Kautz, ``Unsupervised image-to-image translation
  networks,'' in \emph{Proc. NeurIPS}, 2017.

\bibitem{2018-ECCV-Huang-Auggan}
S.-W. Huang, C.-T. Lin, S.-P. Chen, Y.-Y. Wu, P.-H. Hsu, and S.-H. Lai,
  ``Auggan: Cross domain adaptation with gan-based data augmentation,'' in
  \emph{Proc. ECCV}, 2018, pp. 718--731.

\bibitem{2019-WACV-Cherian}
A.~Cherian and A.~Sullivan, ``Sem-gan: semantically-consistent image-to-image
  translation,'' in \emph{Proc. WACV}.\hskip 1em plus 0.5em minus 0.4em\relax
  IEEE, 2019, pp. 1797--1806.

\bibitem{2020-Arxiv-Musto}
L.~Musto and A.~Zinelli, ``Semantically adaptive image-to-image translation for
  domain adaptation of semantic segmentation,'' \emph{arXiv preprint
  arXiv:2009.01166}, 2020.

\bibitem{2020-WACV-Pizzati}
F.~Pizzati, R.~d. Charette, M.~Zaccaria, and P.~Cerri, ``Domain bridge for
  unpaired image-to-image translation and unsupervised domain adaptation,'' in
  \emph{Proc. WACV}, 2020, pp. 2990--2998.

\bibitem{2022-TITS-Luo}
F.~Luo, Y.~Li, G.~Zeng, P.~Peng, G.~Wang, and Y.~Li, ``Thermal infrared image
  colorization for nighttime driving scenes with top-down guided attention,''
  \emph{IEEE Trans. Intell. Transp. Syst.}, 2022.

\bibitem{2021-ICIG-Luo}
F.~Luo, Y.~Cao, and Y.~Li, ``Nighttime thermal infrared image colorization with
  dynamic label mining,'' in \emph{International Conference on Image and
  Graphics}.\hskip 1em plus 0.5em minus 0.4em\relax Springer, 2021, pp.
  388--399.

\bibitem{2012-Gentner}
D.~Gentner and L.~Smith, ``Analogical reasoning. encyclopedia of human
  behavior, 1, 130-136,'' 2012.

\bibitem{2013-Oxford-Holyoak}
K.~J. Holyoak, ``Analogy and relational reasoning,'' \emph{The Oxford Handbook
  of Thinking and Reasoning}, p. 234, 2013.

\bibitem{2019-FLIR-FA}
F.A.Group, ``Flir thermal dataset for algorithm training,''
  \url{https://www.flir.co.uk/oem/adas/adas-dataset-form/}, May 2019.

\bibitem{2015-CVPR-Hwang}
S.~Hwang, J.~Park, N.~Kim, Y.~Choi, and I.~So~Kweon, ``Multispectral pedestrian
  detection: Benchmark dataset and baseline,'' in \emph{Proc. CVPR}, 2015, pp.
  1037--1045.

\bibitem{2018-Berg-CVPRW}
A.~Berg, J.~Ahlberg, and M.~Felsberg, ``Generating visible spectrum images from
  thermal infrared,'' in \emph{Proc. CVPR Workshops}, 2018, pp. 1143--1152.

\bibitem{2020-Bhat-ICCES}
N.~Bhat, N.~Saggu, S.~Kumar \emph{et~al.}, ``Generating visible spectrum images
  from thermal infrared using conditional generative adversarial networks,'' in
  \emph{ICCES}, 2020, pp. 1390--1394.

\bibitem{2020-Kuang-IPT}
X.~Kuang, J.~Zhu, X.~Sui, Y.~Liu, C.~Liu, Q.~Chen, and G.~Gu, ``Thermal
  infrared colorization via conditional generative adversarial network,''
  \emph{Infrared Physics \& Technology}, p. 103338, 2020.

\bibitem{2021-ISEE-Le}
T.~Le-Tien, T.~H.~D. Quang, H.~Y. Vy, T.~Nguyen-Thanh, and H.~Phan-Xuan,
  ``Gan-based thermal infrared image colorization for enhancing object
  identification,'' in \emph{2021 International Symposium on Electrical and
  Electronics Engineering (ISEE)}.\hskip 1em plus 0.5em minus 0.4em\relax IEEE,
  2021, pp. 90--94.

\bibitem{2018-Nyberg-ECCV}
A.~Nyberg, A.~Eldesokey, D.~Bergstrom, and D.~Gustafsson, ``Unpaired thermal to
  visible spectrum transfer using adversarial training,'' in \emph{Proc. ECCV},
  2018, pp. 0--0.

\bibitem{2020-ECCV-Zheng}
Z.~Zheng, Y.~Wu, X.~Han, and J.~Shi, ``Forkgan: Seeing into the rainy night,''
  in \emph{Proc. ECCV}.\hskip 1em plus 0.5em minus 0.4em\relax Springer, 2020,
  pp. 155--170.

\bibitem{2020-CVPR-Chen}
R.~Chen, W.~Huang, B.~Huang, F.~Sun, and B.~Fang, ``Reusing discriminators for
  encoding: Towards unsupervised image-to-image translation,'' in \emph{Proc.
  CVPR}, 2020, pp. 8168--8177.

\bibitem{2018-ECCV-Huang}
X.~Huang, M.-Y. Liu, S.~Belongie, and J.~Kautz, ``Multimodal unsupervised
  image-to-image translation,'' in \emph{Proc. ECCV}, 2018, pp. 172--189.

\bibitem{2020-IJCV-Lee}
H.-Y. Lee, H.-Y. Tseng, Q.~Mao, J.-B. Huang, Y.-D. Lu, M.~Singh, and M.-H.
  Yang, ``Drit++: Diverse image-to-image translation via disentangled
  representations,'' \emph{International Journal of Computer Vision}, vol. 128,
  no.~10, pp. 2402--2417, 2020.

\bibitem{2019-ICRA-Anoosheh}
A.~Anoosheh, T.~Sattler, R.~Timofte, M.~Pollefeys, and L.~Van~Gool,
  ``Night-to-day image translation for retrieval-based localization,'' in
  \emph{ICRA}, 2019, pp. 5958--5964.

\bibitem{2015-ICLR-Weston}
\BIBentryALTinterwordspacing
J.~Weston, S.~Chopra, and A.~Bordes, ``Memory networks,'' in \emph{3rd
  International Conference on Learning Representations, {ICLR} 2015, San Diego,
  CA, USA, May 7-9, 2015, Conference Track Proceedings}, Y.~Bengio and
  Y.~LeCun, Eds., 2015. [Online]. Available:
  \url{http://arxiv.org/abs/1410.3916}
\BIBentrySTDinterwordspacing

\bibitem{2015-NeurIPS-Sukhbaatar}
S.~Sukhbaatar, J.~Weston, R.~Fergus \emph{et~al.}, ``End-to-end memory
  networks,'' \emph{Proc. NeurIPS}, vol.~28, 2015.

\bibitem{2021-ICCV-Xie}
G.-S. Xie, H.~Xiong, J.~Liu, Y.~Yao, and L.~Shao, ``Few-shot semantic
  segmentation with cyclic memory network,'' in \emph{Proc. ICCV}, 2021, pp.
  7293--7302.

\bibitem{2021-ICCV-Wu}
Z.~Wu, X.~Shi, G.~Lin, and J.~Cai, ``Learning meta-class memory for few-shot
  semantic segmentation,'' in \emph{Proc. ICCV}, 2021, pp. 517--526.

\bibitem{2021-ICCV-Alonso}
I.~Alonso, A.~Sabater, D.~Ferstl, L.~Montesano, and A.~C. Murillo,
  ``Semi-supervised semantic segmentation with pixel-level contrastive learning
  from a class-wise memory bank,'' in \emph{Proc. ICCV}, 2021, pp. 8219--8228.

\bibitem{2021-CVPR-VS}
V.~VS, V.~Gupta, P.~Oza, V.~A. Sindagi, and V.~M. Patel, ``Mega-cda: Memory
  guided attention for category-aware unsupervised domain adaptive object
  detection,'' in \emph{Proc. CVPR}, 2021, pp. 4516--4526.

\bibitem{2021-ICCV-Chen}
Y.~Chen, Y.~Wang, Y.~Pan, T.~Yao, X.~Tian, and T.~Mei, ``A style and semantic
  memory mechanism for domain generalization,'' in \emph{Proc. ICCV}, 2021, pp.
  9164--9173.

\bibitem{2021-CVPR-Jeong}
S.~Jeong, Y.~Kim, E.~Lee, and K.~Sohn, ``Memory-guided unsupervised
  image-to-image translation,'' in \emph{Proc. CVPR}, 2021, pp. 6558--6567.

\bibitem{2019-detectron2-Wu}
Y.~Wu, A.~Kirillov, F.~Massa, W.-Y. Lo, and R.~Girshick, ``Detectron2,''
  \url{https://github.com/facebookresearch/detectron2}, 2019.

\bibitem{2020-Arxiv-Tao}
A.~Tao, K.~Sapra, and B.~Catanzaro, ``Hierarchical multi-scale attention for
  semantic segmentation,'' \emph{arXiv preprint arXiv:2005.10821}, 2020.

\bibitem{2018-ECCV-Wu}
Y.~Wu and K.~He, ``Group normalization,'' in \emph{Proc. ECCV}, 2018, pp.
  3--19.

\bibitem{1992-PD-Rudin}
L.~I. Rudin, S.~Osher, and E.~Fatemi, ``Nonlinear total variation based noise
  removal algorithms,'' \emph{Physica D: nonlinear phenomena}, vol.~60, no.
  1-4, pp. 259--268, 1992.

\bibitem{2004-TIP-Wang}
Z.~Wang, A.~C. Bovik, H.~R. Sheikh, and E.~P. Simoncelli, ``Image quality
  assessment: from error visibility to structural similarity,'' \emph{IEEE
  Trans. Image Process.}, vol.~13, no.~4, pp. 600--612, 2004.

\bibitem{2018-Arxiv-Miyato}
T.~Miyato, T.~Kataoka, M.~Koyama, and Y.~Yoshida, ``Spectral normalization for
  generative adversarial networks,'' \emph{arXiv preprint arXiv:1802.05957},
  2018.

\bibitem{2018-Arxiv-Jolicoeur}
A.~Jolicoeur-Martineau, ``The relativistic discriminator: a key element missing
  from standard gan,'' \emph{arXiv preprint arXiv:1807.00734}, 2018.

\bibitem{2019-TVT-Ma}
Z.~Ma, D.~Chang, J.~Xie, Y.~Ding, S.~Wen, X.~Li, Z.~Si, and J.~Guo,
  ``Fine-grained vehicle classification with channel max pooling modified
  cnns,'' \emph{IEEE Trans. Veh. Technol.}, vol.~68, no.~4, pp. 3224--3233,
  2019.

\bibitem{2014-ECCV-Lin}
T.-Y. Lin, M.~Maire, S.~Belongie, J.~Hays, P.~Perona, D.~Ramanan,
  P.~Doll{\'a}r, and C.~L. Zitnick, ``Microsoft coco: Common objects in
  context,'' in \emph{Proc. ECCV}, 2014, pp. 740--755.

\bibitem{2016-CVPR-Cordts}
M.~Cordts, M.~Omran, S.~Ramos, T.~Rehfeld, M.~Enzweiler, R.~Benenson,
  U.~Franke, S.~Roth, and B.~Schiele, ``The cityscapes dataset for semantic
  urban scene understanding,'' in \emph{Proc. CVPR}, 2016, pp. 3213--3223.

\bibitem{2019-IJCAI-Zheng}
Z.~Zheng and Y.~Yang, ``Unsupervised scene adaptation with memory
  regularization in vivo,'' in \emph{Proc. IJCAI}, 2020.

\bibitem{2020-CVPR-Zhen}
M.~Zhen, J.~Wang, L.~Zhou, S.~Li, T.~Shen, J.~Shang, T.~Fang, and L.~Quan,
  ``Joint semantic segmentation and boundary detection using iterative pyramid
  contexts,'' in \emph{Proc. CVPR}, 2020, pp. 13\,666--13\,675.

\bibitem{1967-BSM-MacQueen}
J.~MacQueen, ``Classification and analysis of multivariate observations,'' in
  \emph{5th Berkeley Symp. Math. Statist. Probability}, 1967, pp. 281--297.

\bibitem{1982-TIT-Lloyd}
S.~Lloyd, ``Least squares quantization in pcm,'' \emph{IEEE Trans. Inf.
  Theory}, vol.~28, no.~2, pp. 129--137, 1982.

\bibitem{2020-ICML-Chen}
T.~Chen, S.~Kornblith, M.~Norouzi, and G.~Hinton, ``A simple framework for
  contrastive learning of visual representations,'' in \emph{International
  conference on machine learning}.\hskip 1em plus 0.5em minus 0.4em\relax PMLR,
  2020, pp. 1597--1607.

\bibitem{2020-TNNLS-Zhao}
C.~Zhao, G.~G. Yen, Q.~Sun, C.~Zhang, and Y.~Tang, ``Masked gan for
  unsupervised depth and pose prediction with scale consistency,'' \emph{IEEE
  Trans. Neural Netw. Learn. Syst.}, vol.~32, no.~12, pp. 5392--5403, 2020.

\bibitem{2015-IJCV-Everingham}
M.~Everingham, S.~A. Eslami, L.~Van~Gool, C.~K. Williams, J.~Winn, and
  A.~Zisserman, ``The pascal visual object classes challenge: A
  retrospective,'' \emph{International journal of computer vision}, vol. 111,
  no.~1, pp. 98--136, 2015.

\bibitem{2010-IJCV-Everingham}
M.~Everingham, L.~Van~Gool, C.~K. Williams, J.~Winn, and A.~Zisserman, ``The
  pascal visual object classes (voc) challenge,'' \emph{International journal
  of computer vision}, vol.~88, no.~2, pp. 303--338, 2010.

\bibitem{2014-Arxiv-Kingma}
D.~P. Kingma and J.~Ba, ``Adam: A method for stochastic optimization,'' in
  \emph{ICLR (Poster)}, 2015.

\bibitem{2020-Arxiv-Bochkovskiy}
A.~Bochkovskiy, C.-Y. Wang, and H.-Y.~M. Liao, ``Yolov4: Optimal speed and
  accuracy of object detection,'' \emph{arXiv preprint arXiv:2004.10934}, 2020.

\end{thebibliography}

\begin{IEEEbiography}{Fu-Ya Luo}
received the B.S. degree in biomedical engineering from University of Electronic Science 
and Technology of China (UESTC), in 2015.
He is now pursuing his Ph.D. degree in UESTC.
His research interests include scene understanding, brain-inspired computer vision, 
weakly supervised learning, and image-to-image translation.
\end{IEEEbiography}

\begin{IEEEbiography}{Yi-Jun Cao}
received the M.S. degree from the College of Electric and Information Engineering,
Guangxi University of Science and Technology. Currently, he is working toward the Ph.D. degree 
with the College of Electric and Information Engineering, University of Electronic Science and 
Technology of China (UESTC). His area of research are visual SLAM and navigation.
\end{IEEEbiography}

\begin{IEEEbiography}{Kai-Fu Yang}
received the Ph.D. degree in biomedical engineering from the University of Electronic Science 
and Technology of China (UESTC), Chengdu, China, in 2016. He is currently an associate research 
professor with the MOE Key Lab for Neuroinformation, School of Life Science and Technology, 
UESTC, Chengdu, China. His research interests include cognitive computing and braininspired 
computer vision.
\end{IEEEbiography}

\begin{IEEEbiography}{Yong-Jie Li}
(Senior Member, IEEE) received the Ph.D. degree in biomedical engineering from
University of Electronic Science and Technology of China (UESTC), in 2004. He is currently a Professor 
with the Key Laboratory for NeuroInformation of Ministry of Education, School of Life 
Science and Technology, UESTC. His research focuses on building of biologically inspired 
computational models of visual perception and the applications in image processing and 
computer vision.
\end{IEEEbiography}

\end{document}